\documentclass{article}

\usepackage{microtype}
\usepackage{graphicx}
\usepackage{subcaption}
\usepackage{booktabs} % for professional tables

\usepackage{hyperref}

\usepackage[preprint]{icml2026}

\usepackage{amsmath}
\usepackage{amssymb}
\usepackage{mathtools}
\usepackage{amsthm}

\usepackage{float}
\usepackage{stfloats}

\usepackage{multirow}
\usepackage{inconsolata}
\usepackage{makecell}
\usepackage{titlesec}

\usepackage{pdfpages}
\usepackage[capitalize,noabbrev]{cleveref}

\theoremstyle{plain}

\theoremstyle{definition}

\theoremstyle{remark}

\usepackage[textsize=tiny]{todonotes}

\icmltitlerunning{Vectra: Visual Quality Assessment for E-Commerce In-Image Machine Translation}

\titleformat{\paragraph}[runin]{\normalfont\bfseries}{}{0em}{}

\titlespacing*{\paragraph}{0pt}{0pt}{1em}

\begin{document}

\twocolumn[
  \icmltitle{Vectra: A New Metric, Dataset, and Model for Visual Quality Assessment\\in E-Commerce In-Image Machine Translation}



  \icmlsetsymbol{equal}{*}

  \begin{icmlauthorlist}
    \icmlauthor{Qingyu Wu}{equal,ali}
    \icmlauthor{Yuxuan Han}{equal,ali}
    \icmlauthor{Haijun Li}{ali}
    \icmlauthor{Zhao Xu}{ali}
    \icmlauthor{Jianshan Zhao}{ali}
    \icmlauthor{Xu Jin}{ali}
    \icmlauthor{Longyue Wang}{ali}
    \icmlauthor{Weihua Luo}{ali}
  \end{icmlauthorlist}

  \icmlaffiliation{ali}{Alibaba International Digital Commerce Group}

  \icmlcorrespondingauthor{Zhao Xu}{changgong.xz@alibaba-inc.com}

  \icmlkeywords{In-Image Machine Translation, Visual Quality Assessment, Evaluation Metrics, Multimodal Large Language Models, Datasets and Benchmarks}

  \vskip 0.3in
]



%\printAffiliationsAndNotice{}  % no special notice (required even if empty)
\printAffiliationsAndNotice{\icmlEqualContribution}

%Title: Vectra: A New Metric, Dataset, and Model for Visual Quality Assessment\\in E-commerce In-Image Machine Translation
% ... existing code ...

\begin{abstract}
In-Image Machine Translation (IIMT) powers cross-border e-commerce product listings; existing research focuses on machine translation evaluation, while visual rendering quality is critical for user engagement. When facing context-dense product imagery and multimodal defects, current reference-based methods (e.g., SSIM, FID) lack explainability, while model-as-judge approaches lack domain-grounded, fine-grained reward signals. To bridge this gap, we introduce Vectra, to the best of our knowledge, the first reference-free, MLLM-driven visual quality assessment framework for e-commerce IIMT. Vectra comprises three components: (1) Vectra Score, a multidimensional quality metric system that decomposes visual quality into 14 interpretable dimensions, with spatially-aware Defect Area Ratio (DAR) quantification to reduce annotation ambiguity; (2) Vectra Dataset, constructed from 1.1M real-world product images via diversity-aware sampling, comprising a 2K benchmark for system evaluation, 30K reasoning-based annotations for instruction tuning, and 3.5K expert-labeled preferences for alignment and evaluation; and (3) Vectra Model, a 4B-parameter MLLM that generates both quantitative scores and diagnostic reasoning. Experiments demonstrate that Vectra achieves state-of-the-art correlation with human rankings, and our model outperforms leading MLLMs, including GPT-5 and Gemini-3, in scoring performance. The dataset and model will be released upon acceptance.
\end{abstract}
\vspace{-0.8cm}
%Beyond e-commerce, Vectra's interpretable evaluation paradigm offers a generalizable framework for assessing multimodal generation quality.

\begin{figure}[!t]
  \centering
\includegraphics[width=0.87\columnwidth]{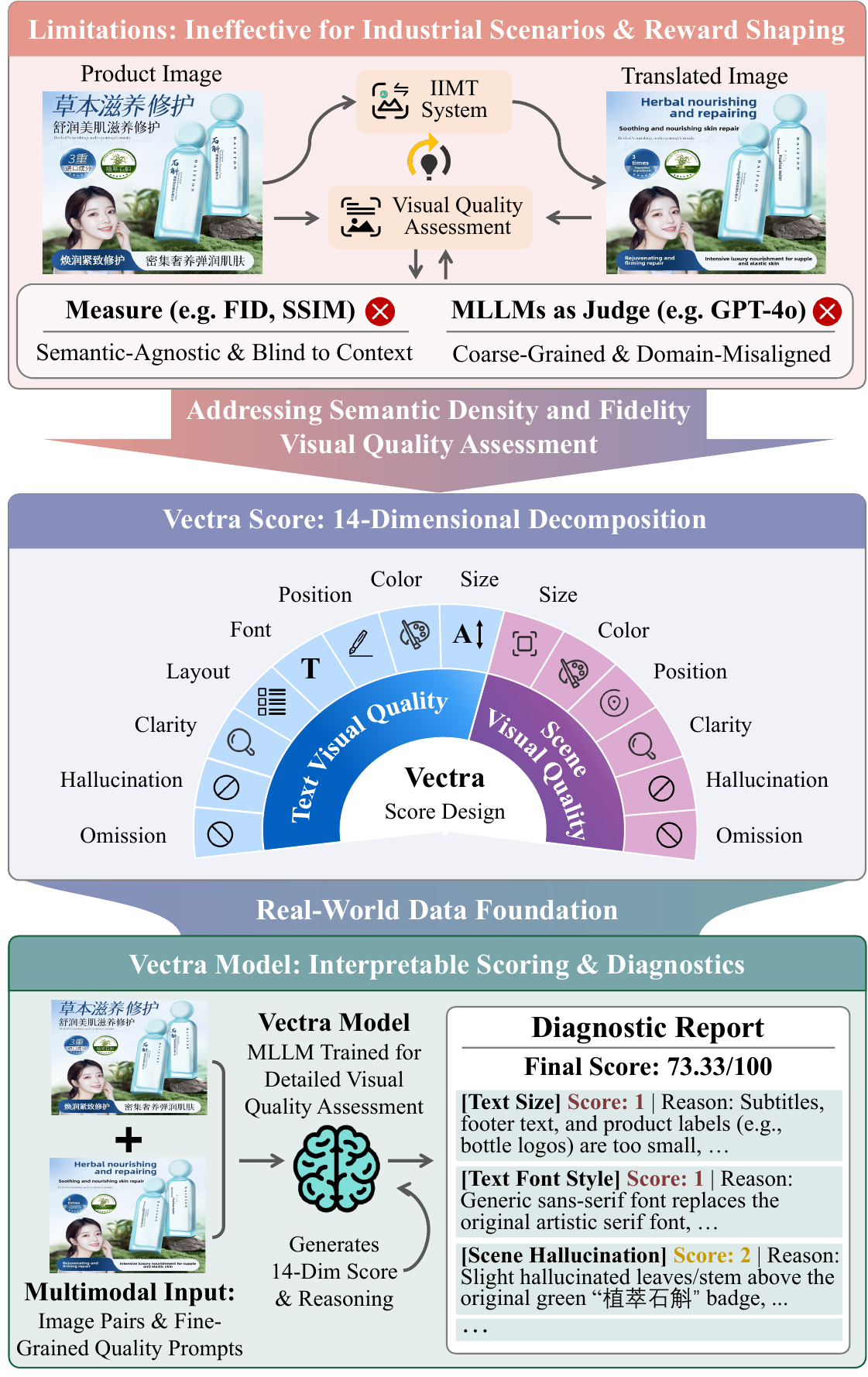}
\vspace{-0.1cm}
\caption{An illustration of the limitations in current IIMT visual quality assessment. Existing methods (Top) struggle to pinpoint fine-grained defects or serve as effective reward signals in contextually dense e-commerce scenarios. In contrast, Vectra decomposes visual quality into 14 dimensions (Middle), enabling precise error detection and reasoning-based diagnostics (Bottom).}
  \label{fig:introduction}
  \vspace{-0.6cm}
\end{figure}

\section{Introduction}
In-Image Machine Translation (IIMT) extends traditional Text Image Translation (TIT) by requiring not only translating textual content but also preserving the visual consistency of both text and scene from the source image. With advances in neural networks, research has increasingly shifted toward end-to-end systems that jointly handle text detection, translation, and rendering \cite{IIMT-1,lantranslatotron,tian2025prim}. Due to the multimodal nature of IIMT, evaluation has traditionally combined machine translation metrics with visual quality assessment. Early work, with modest visual fidelity requirements, relied on reference-based methods such as SSIM \cite{SSIM} and FID \cite{FID} that compare outputs against gold-standard references. With the development of MLLM and the growing demands for fine-grained visual quality (e.g., font style, layout consistency), researchers turned to model-as-judge approaches based on MLLM \cite{fu2024ensuring}.

However, the rapid growth of cross-border e-commerce has fundamentally raised the stakes. LLM/MLLM-powered IIMT systems now drive global product listings \cite{AIDC2025,GhostCut2025}, creating urgent demand for open-set (no reference), fine-grained visual quality feedback. IIMT evaluation requires precise comparison of multimodal defects (textual and non-textual) across source-target image pairs, yet even leading MLLMs exhibit significant limitations in providing fine-grained, interpretable assessments (as shown in Appendix~\ref{app:reason_quality}) and lack domain-specific calibration. Drawing inspiration from the evolution of MT evaluation, which progressed from formula-based metrics to embedding-based approaches like COMET \cite{COMET}, and further to LLM-driven interpretable evaluation like InstructScore \cite{INSTRUCTSCORE}, we propose an MLLM-based framework that delivers interpretable, domain-efficient, fine-grained visual quality assessment for IIMT, as illustrated in Figure~\ref{fig:introduction}.

To address these challenges, we introduce \textbf{Vectra}, a comprehensive evaluation framework comprising three components (Figure~\ref{fig:vectra_framework}): (1) \textbf{Vectra Score}, a multidimensional metric decomposing visual quality into 14 interpretable dimensions with Defect Area Ratio (DAR) for spatial severity quantification, reducing inter-annotator Coefficient of Variation (CV) by 46\%; (2) \textbf{Vectra Dataset}, constructed through diversity-aware sampling from 1.1M real-world e-commerce images while ensuring regulatory compliance and distributional coverage. This dataset includes (i) 2K images for Vectra-Bench benchmark evaluation, and (ii) approximately 33.5K images for MLLM-as-judge training and evaluation, combining scalable model predictions with expert annotations from multi-annotator teams across 5 language pairs and synthesized through multiple IIMT systems to capture diverse defect patterns; (3) \textbf{Vectra Model}, a 4B-parameter MLLM fine-tuned on instruction-following data distilled from Gemini-2.5-Pro and aligned with expert preferences, capable of generating both quantitative scores and diagnostic reasoning.

Our contributions are summarized as follows:

\textbullet~To the best of our knowledge, we establish the first systematic framework for e-commerce IIMT visual quality assessment, introducing Vectra Score—a reference-free metric decomposing quality into 14 interpretable dimensions with spatial severity quantification.

\textbullet~We construct a comprehensive dataset ecosystem including 2K benchmark images and 33.5K model-as-judge construction samples from real-world e-commerce scenarios, enabling both model development and standardized evaluation.

\textbullet~We demonstrate the practical viability through Vectra Model, a 4B-parameter MLLM that achieves competitive performance (Pearson $r = 0.895$, Kendall $\tau = 0.724$) with leading MLLMs, provides interpretable diagnostic reasoning, and generalizes well to out-of-domain IIMT scenarios.

\section{The Vectra Evaluation Framework}

\subsection{Metric}
\subsubsection{Quality Dimensions}
\label{sec:taxonomy}
We develop a multidimensional scoring protocol tailored for the spatial and multimodal nature of IIMT visual quality assessment. The evaluation focuses on visual rendering quality of \textit{text regions} (translated textual content) and \textit{scene regions} (non-textual content such as backgrounds, graphics, and product depictions).

\begin{figure*}[ht]
  \includegraphics[width=\textwidth]{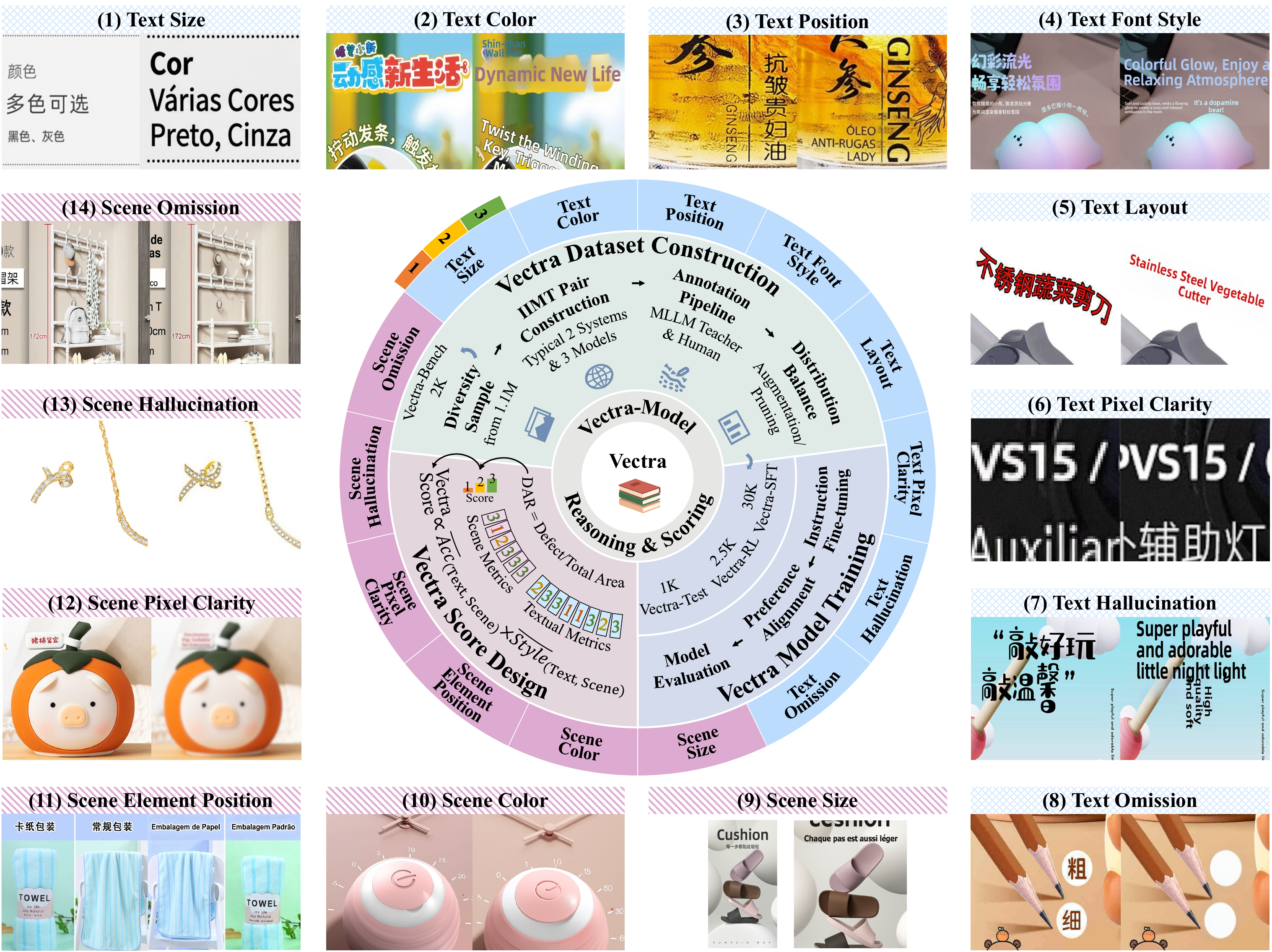}
  \caption{Overview of the \textbf{Vectra framework}. Visual quality is decomposed into 14 dimensions: \textbf{Text} (Blue, 1--8) and \textbf{Scene} (Pink, 9--14), with surrounding examples illustrating representative defects. The center shows the \textbf{Vectra Score} computation via multiplicative aggregation, \textbf{Data Suite} construction, and \textbf{Vectra Model} training pipeline.}
  \label{fig:vectra_framework}
  \vspace{-0.4cm}
\end{figure*}

Traditional multidimensional quality metric frameworks \cite{MQM,5-1}, which are widely used for textual tasks, typically partition error types into \textit{Accuracy}, \textit{Fluency}, \textit{Style}, etc. However, these frameworks focus primarily on linguistic phenomena and do not directly transfer to our work where spatial layout, graphical fidelity, and scene coherence are paramount. Drawing inspiration from MQM-like works, we propose 14 visual quality assessment dimensions (Figure~\ref{fig:vectra_framework}), organized into \textit{Accuracy} (e.g., hallucinations, omissions) and \textit{Style} (e.g., font style, color) categories—where \textit{Style} subsumes fluency-related visual attributes such as layout consistency and spatial alignment. Eight dimensions evaluate \textbf{Textual Visual Quality}, assessing visual artifacts in rendered text. Six dimensions assess \textbf{Scene Visual Quality}, measuring defects in non-textual areas. Detailed dimension definitions are provided in Appendix~\ref{app:scoring_protocol}.

\begin{figure}[htbp]
  \centering
  \includegraphics[width=\columnwidth]{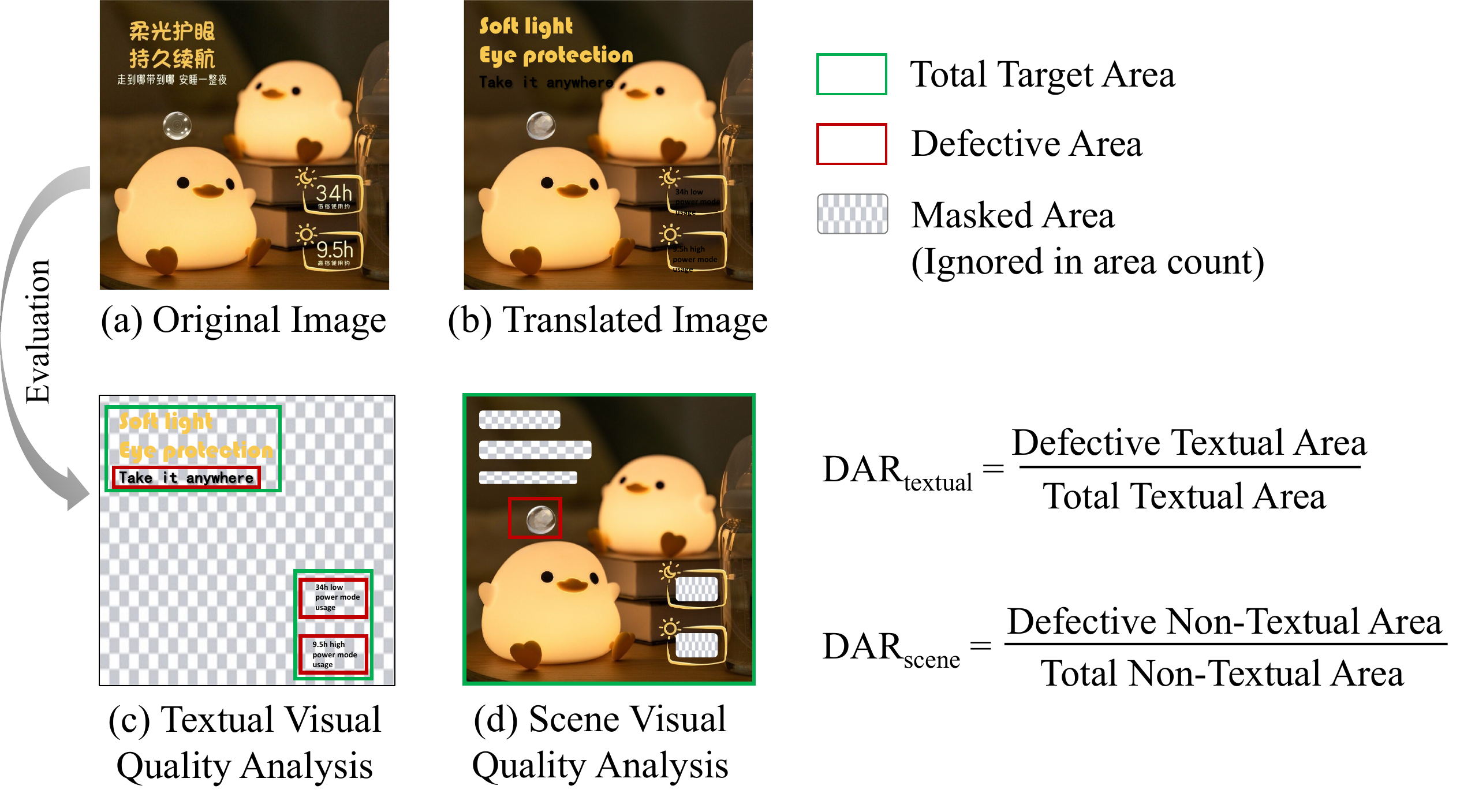}
  \vspace{-0.6cm}
  \caption{Illustration of the Defect Area Ratio (DAR) calculation mechanism between the original image (a) and the translated image (b). The metric quantifies quality by measuring the defective areas (red) relative to the total target areas (green) for both textual (c) and non-textual (d) elements.}
  \label{fig:dar_calculation}
  \vspace{-0.3cm}
\end{figure}

\subsubsection{Defect Quantification}
\label{sec:dar}
To objectively quantify visual defects, we introduce the \textbf{Defect Area Ratio (DAR)}, a spatial metric measuring defect prominence through area coverage. As illustrated in Figure~\ref{fig:dar_calculation}, DAR computes the ratio of defect area to total content area, in which textual and non-textual regions are defined separately.

Following strategies in related works \cite{3-score-1,fu2024ensuring}, we discretize DAR into a 3-point ordinal scale $s \in \{1, 2, 3\}$:
\begin{equation}
    s = \begin{cases} 
    3 \text{ (Excellent)} & \text{if } \text{DAR} \approx 0 \\
    2 \text{ (Fair)} & \text{if } 0 < \text{DAR} \le \tau \\
    1 \text{ (Poor)} & \text{if } \text{DAR} > \tau
    \end{cases}.
\end{equation}

\textbf{Threshold Calibration.} To determine the threshold $\tau$ empirically, we iteratively sampled image pairs from the Vectra Dataset. Using Gemini-2.5-Pro, we identify samples containing one of the error types and estimate their DAR to populate 10 bins ($[0, 0.1), [0.1, 0.2), \dots, [0.9, 1.0]$) with manual verification (40 samples/bin). Five e-commerce experts then evaluate each image via a multiple-choice protocol (blind to DAR values), with the final ground truth established by majority vote.

\begin{figure}[htbp]
\vspace{-0.2cm}
  \centering
  \includegraphics[width=\columnwidth]{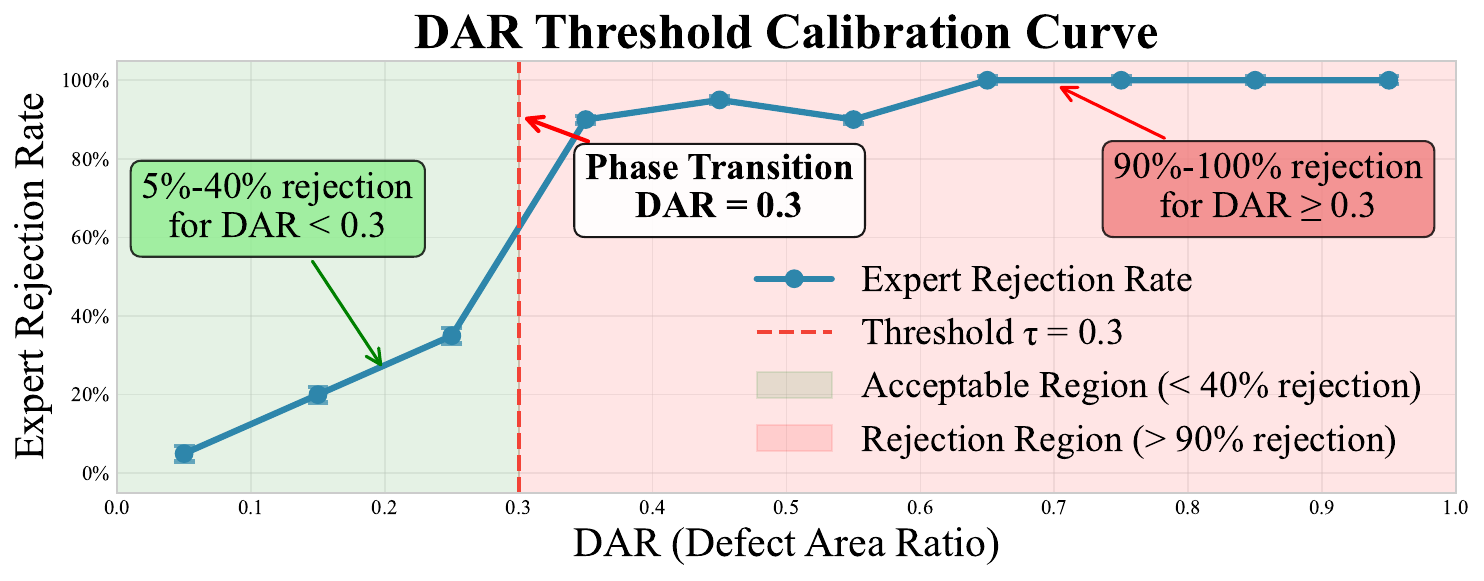}
  \vspace{-0.7cm}
  \caption{Expert rejection rates as a function of DAR values. A sharp increase in rejection rate occurs at $\tau = 0.3$, which we adopt as the DAR threshold for distinguishing Fair (Score 2) from Poor (Score 1) quality.}
  \label{fig:dar_curve}
  \vspace{-0.2cm}
\end{figure}

Figure~\ref{fig:dar_curve} reveals a distinct phase transition: rejection rates remain below 40\% for $\text{DAR} < 0.3$, then surge beyond 90\% for the $[0.3, 0.4)$ bin. This inflection point indicates that defects exceeding 30\% of valid regions consistently violate commercial standards. We therefore set $\tau = 0.3$.

\begin{figure*}[hb]
\vspace{-0.3cm}
  \includegraphics[width=\textwidth]{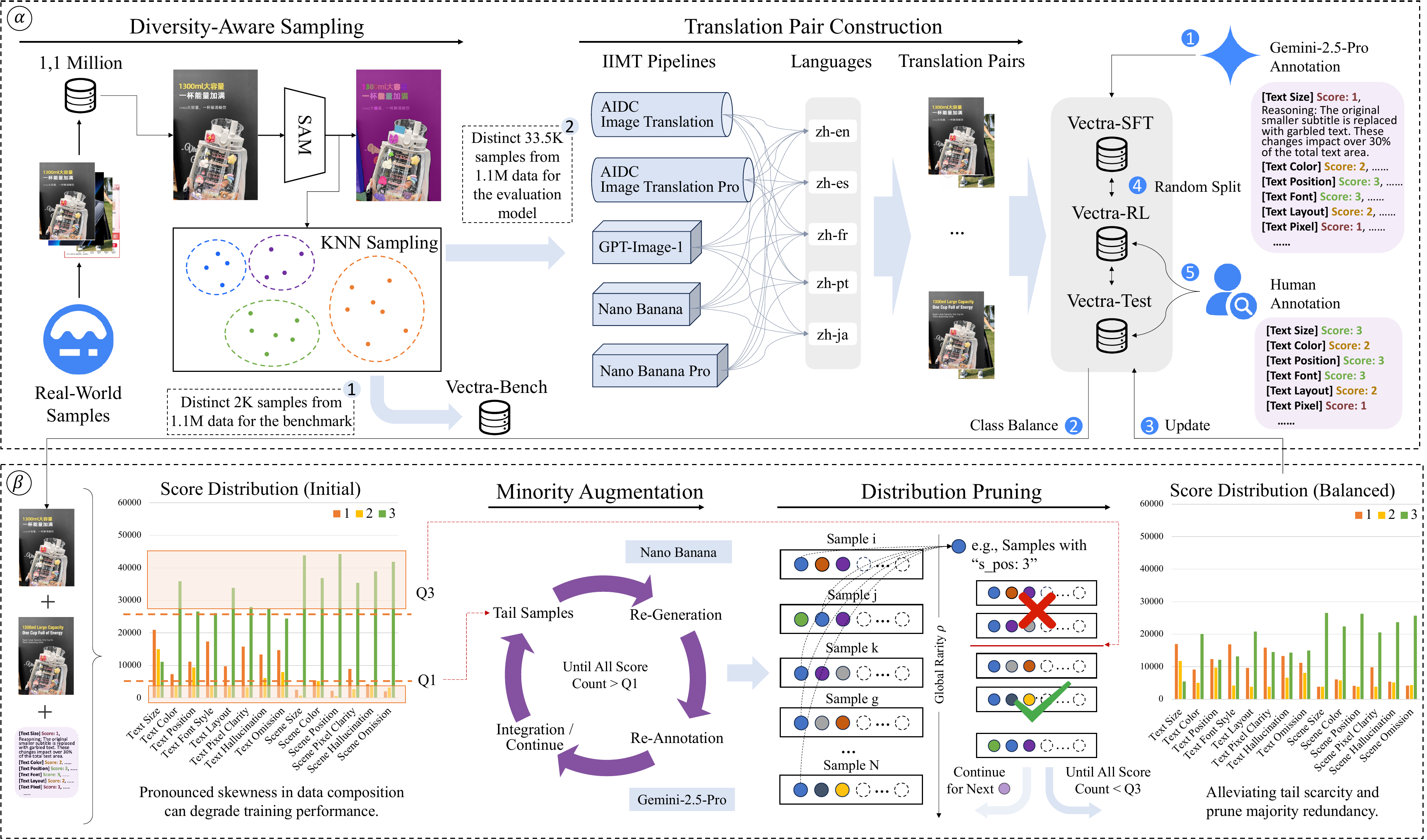}
  \caption{Data pipeline: ($\alpha$) Diversity-aware sampling and translation pair construction; ($\beta$) Distribution balancing via minority augmentation and pruning.}
  \label{fig:vectra_data}
\vspace{-0.3cm}
\end{figure*}

\subsubsection{Overall Score Computation}

In e-commerce scenarios, accuracy errors (e.g., hallucinations, omissions) constitute critical violations that cannot be offset by stylistic excellence—a single mistranslated or fabricated product feature may breach consumer protection regulations.

To encode this non-compensatory relationship, we adopt a multiplicative formulation following established practices in composite indicator design rather than additive averaging \cite{multi}. Let $\bar{s}_{\text{acc}}^{(j)}$ and $\bar{s}_{\text{sty}}^{(j)}$ denote the mean scores across accuracy and style dimensions for sample $j$, and let $\phi(\cdot)$ denote min-max normalization mapping scores from $[s_{\min}, s_{\max}]$ to $[0, 1]$. The final Vectra Score is:
\begin{equation}
\label{eq:vectra_score}
\text{Vectra Score} = 100 \Big[ \phi(\bar{s}_{\text{acc}}^{(j)}) \cdot \phi(\bar{s}_{\text{sty}}^{(j)}) \Big],
\end{equation}
where $s_{\min}=1$ and $s_{\max}=3$. This multiplicative scheme ensures that severe accuracy failures drive the overall score toward zero regardless of style performance, rigorously enforcing the business-critical constraint that content fidelity cannot be traded off against visual aesthetics.

\subsection{Data and Evaluation Model}
\label{sec:evaluation_methodology}
With the Vectra Score established, we now describe the construction of the supporting dataset and the training of a specialized model for automated evaluation.

\subsubsection{Data Collection}

\paragraph{Diversity-Aware Sampling.}
\label{sec:diversity_sampling}
Real-world e-commerce product images exhibit unique redundancy patterns: merchants typically prepare multiple images as a set for each product, and similar products often share nearly identical visual layouts. To maximize semantic diversity from such structured imagery, we employ SAM \cite{SAM-huge} with ViT-H backbone as a feature extractor, capturing both object-level semantics and spatial layout patterns. Operating on 1.1M real-world product images from consenting merchants under data protection compliance, we perform k-means clustering in the SAM feature space and select medoid images from each cluster as illustrated in Figure~\ref{fig:vectra_data} ($\alpha$). This yields Vectra-Bench comprising 2,000 maximally diverse images for evaluation, plus 33.5K additional samples for Vectra Model training and testing (see Appendix~\ref{app:data_sampling} for details).

\paragraph{Translation Pair Construction.} 
To ensure comprehensive coverage of quality distributions and architectural diversity, we leverage five translation engines: AIDC Image Translation Pro/Standard (e-commerce solutions), and three end-to-end models (GPT-Image-1, Nano Banana Pro, Nano Banana). Each engine generates outputs for five language pairs ($zh \rightarrow \{en, es, fr, pt, ja\}$) as illustrated in Figure~\ref{fig:vectra_data}($\alpha$). Prompting strategies are detailed in Appendix~\ref{app:translation_prompt}.

\subsubsection{Distribution Balancing}
We balance the marginal distribution of 42 dimension-score pairs (14 dimensions × 3 levels), rather than the intractable $3^{14}$ joint combinations. This enables controllable synthesis of specific error types with quality verification while preserving sensitivity to rare patterns in each dimension. The pipeline operates in two phases as illustrated in Figure~\ref{fig:vectra_data} ($\beta$) and Algorithm~\ref{alg:distribution_balancing}, we adopt quartile thresholds $Q_1$ (25th percentile) and $Q_3$ (75th percentile), a common imbalance ratio \cite{Q13,Q13-2}. \textbf{Phase 1} augments scarce dimension-score tuples below $Q_1$ by synthesizing candidates via Nano Banana, with Gemini-2.5-Pro verification ensuring the target defect is present (see Appendix~\ref{app:class_balance} for augmentation examples). \textbf{Phase 2} prunes overrepresented tuples above $Q_3$ by removing samples with lowest global rarity score $\gamma(x) = \sum_{d=1}^{14} \mathcal{F}_{d,x_d}^{-1}$.

\begin{algorithm}[htbp]
\caption{Distribution Balancing}
\footnotesize
\label{alg:distribution_balancing}
\begin{algorithmic}[1]
\REQUIRE Dataset $\mathcal{D}$ with $N$ samples, each sample $x$ has scores $x_d \in \{1,2,3\}$ for dimension $d \in \{1,...,14\}$
\ENSURE Balanced dataset $\mathcal{D}'$
\STATE \textit{// Compute marginal frequencies and thresholds}
\STATE $\mathcal{F}_{d,s} \gets |\{x \in \mathcal{D} : x_d = s\}|$ for all dimension-score pairs $(d,s)$
\STATE $\tau_{\text{low}} \gets Q_1(\{\mathcal{F}_{d,s}\}_{d,s})$; \ $\tau_{\text{high}} \gets Q_3(\{\mathcal{F}_{d,s}\}_{d,s})$
\STATE \textit{// Phase 1: Minority Augmentation}
\FOR{each $(d,s)$ where $\mathcal{F}_{d,s} < \tau_{\text{low}}$}
    \WHILE{$\mathcal{F}_{d,s} < \tau_{\text{low}}$}
        \STATE \textit{// Generation via Nano Banana}
        \STATE $x_{\text{new}} \gets \textsc{Synthesize}(d, s)$
        \STATE \textit{// Verification via Gemini-2.5-Pro}
        \IF{$\textsc{Verify}(x_{\text{new}})$ returns $x_{\text{new},d} = s$}
            \STATE $\mathcal{D} \gets \mathcal{D} \cup \{x_{\text{new}}\}$; \ $\mathcal{F}_{d,s} \gets \mathcal{F}_{d,s} + 1$
        \ENDIF
    \ENDWHILE
\ENDFOR
\STATE \textit{// Phase 2: Distribution Pruning}
\WHILE{$\exists (d,s): \mathcal{F}_{d,s} > \tau_{\text{high}}$}
    \STATE \textit{// Find most overrepresented pair and global rarity}
    \STATE $(d^*, s^*) \gets \arg\max_{(d,s)} (\mathcal{F}_{d,s} - \tau_{\text{high}})$
    \STATE $\gamma(x) \gets \sum_{d=1}^{14} \mathcal{F}_{d,x_d}^{-1}$ for all $x \in \mathcal{D}$
    \STATE $\mathcal{C} \gets \{x \in \mathcal{D} : x_{d^*} = s^*\}$; \ sort $\mathcal{C}$ by $\gamma(x)$ ascending
    \STATE $n \gets \mathcal{F}_{d^*,s^*} - \tau_{\text{high}}$; \ remove first $n$ samples from $\mathcal{C}$
    \STATE $\mathcal{D} \gets \mathcal{D} \setminus \{\text{removed samples}\}$; \ update $\mathcal{F}$
\ENDWHILE
\STATE \textbf{return} $\mathcal{D}$
\end{algorithmic}
\end{algorithm}

\subsubsection{Vectra Model}
\label{sec:vectra_model}
To balance computational bottleneck with open-set evaluation scalability, we fine-tune Qwen3-VL-4B-Instruct \cite{Qwen3VL} through two-stage training: supervised fine-tuning followed by preference alignment.

\paragraph{Supervised Fine-tuning.}
The model is fine-tuned on Vectra-SFT using standard SFT loss. Each sample consists of an image pair (the original source and the translated image), an instruction input (detailed in Appendix~\ref{app:scoring_protocol}), and a structured response. To ensure interpretability and precise error localization, we enforce a schema-guided reasoning format. Specifically, for each dimension, the responses are trying to distill a rationale chain from the large-scale teacher MLLM that follows a logic: identifying the \texttt{CONTENT}, describing the \texttt{ISSUE}, localizing the \texttt{POSITION}, and quantifying the \texttt{EFFECT} (Defect Area Ratio), prior to assigning a score. The output is wrapped in XML tags corresponding to each of the 14 dimensions to facilitate parsing (e.g., \texttt{<t\_size\_reason>...\allowbreak</t\_size\_reason>\allowbreak<t\_size\_score>...\allowbreak</t\_size\_score>}).

\begin{table*}[hb]
\vspace{-0.3cm}
\centering
\scriptsize
\renewcommand{\arraystretch}{0.9}
\setlength{\tabcolsep}{3pt}
\caption{Evaluation of metric performance via Pearson ($r$) and Kendall ($\tau$) correlations, reflecting the consistency between the system rankings calculated by each metric and the human ground-truth rankings on Vectra and OOD benchmarks.}
\label{tab:sample_correlation}
\resizebox{\textwidth}{!}{%
\begin{tabular}{lcccccccccc}
\toprule
 & \multicolumn{6}{c}{\textbf{Vectra} ($r$/$\tau$)} & \multicolumn{4}{c}{\textbf{OOD} ($r$/$\tau$)} \\
\cmidrule(lr){2-7} \cmidrule(lr){8-11}
\textbf{Metric} & zh$\rightarrow$en & zh$\rightarrow$es & zh$\rightarrow$pt & zh$\rightarrow$ja & zh$\rightarrow$fr & Overall & Document & Poster & Scene & Overall \\
\midrule
\multicolumn{11}{l}{\textit{Reference-based Metrics (Top-Ranked Candidate as Reference)}} \\
SSIM & 0.567/0.448 & 0.432/0.337 & 0.586/0.476 & 0.537/0.438 & 0.447/0.359 & 0.514/0.412 & 0.157/0.119 & 0.532/0.414 & \textbf{0.552/0.448} & 0.400/0.318 \\
PSNR & 0.537/0.434 & 0.374/0.298 & 0.590/0.485 & 0.525/0.431 & 0.443/0.364 & 0.494/0.403 & -0.136/-0.085 & 0.501/0.422 & 0.508/0.406 & 0.264/0.224 \\
LPIPS & 0.303/0.237 & 0.231/0.183 & 0.368/0.296 & 0.331/0.268 & 0.262/0.215 & 0.299/0.240 & 0.077/0.059 & 0.180/0.158 & 0.448/0.383 & 0.261/0.221 \\
FID & 0.660/0.557 & 0.586/0.483 & 0.695/0.599 & 0.559/0.465 & 0.618/0.516 & 0.624/0.525 & 0.204/0.170 & 0.288/0.202 & 0.521/0.414 & 0.361/0.285 \\
\multicolumn{11}{l}{\textit{No-Reference Metrics (Direct Assessment)}} \\
SAMScore & 0.337/0.231 & 0.122/0.096 & 0.364/0.257 & 0.401/0.306 & 0.234/0.166 & 0.292/0.211 & 0.156/0.122 & 0.318/0.232 & 0.315/0.241 & 0.256/0.195 \\
E-Commerce\_VQA & 0.047/0.031 & 0.008/0.001 & 0.044/0.025 & -0.080/-0.068 & 0.093/0.074 & 0.025/0.014 & -0.043/-0.036 & -0.094/-0.083 & -0.159/-0.116 & -0.050/-0.041 \\
HCIIT-Judge & 0.602/0.496 & 0.577/0.466 & 0.647/0.544 & 0.538/0.460 & 0.627/0.541 & 0.599/0.502 & 0.553/0.333 & 0.278/0.067 & 0.448/0.067 & 0.435/0.200 \\
\textbf{Vectra} & \textbf{0.800/0.709} & \textbf{0.732/0.588} & \textbf{0.773/0.675} & \textbf{0.677/0.599} & \textbf{0.708/0.613} & \textbf{0.738/0.637} & \textbf{0.599/0.490} & \textbf{0.631/0.523} & 0.496/0.423 & \textbf{0.558/0.466} \\
\bottomrule
\end{tabular}%
}
\vspace{-0.3cm}
\end{table*}

\paragraph{Preference Alignment.}
To align with human judgment, we perform reinforcement learning on 2K expert-annotated samples (Vectra-RL), where each image is evaluated by 5 e-commerce experts. We adopt Group Sequence Policy Optimization (GSPO) \cite{GSPO} for its training stability:

\begin{equation}
\begin{aligned}
\mathcal{J}_{\text{GSPO}}(\theta) &= \mathbb{E}_{x \sim \mathcal{D}, \{y_i\}_{i=1}^G \sim \pi_{\text{old}}} \bigg[ \frac{1}{G} \sum_{i=1}^{G} \min \big(\\
&\quad s_i(\theta)\hat{A}_i, \text{clip}(s_i(\theta), 1-\epsilon, 1+\epsilon)\hat{A}_i \big) \bigg],
\end{aligned}
\end{equation}
where $G$ is the group size, $s_i(\theta) = \left(\frac{\pi_\theta(y_i|x)}{\pi_{\theta_{\text{old}}}(y_i|x)}\right)^{\frac{1}{|y_i|}}$ is the length-normalized importance ratio, and $\hat{A}_i$ is the standardized reward advantage within each response group. The reward function combines two components: $r(y, y^*) = r_{\text{format}}(y) + r_{\text{preference}}(y, y^*)$, where $y$ is the model output and $y^*$ is the expert annotation. Specifically, $r_{\text{format}}(y)$ ensures the output adheres to the required template structure, and $r_{\text{preference}}(y, y^*)$ measures the alignment between model predictions and expert judgments. Implementation details are provided in Appendix~\ref{app:rl_config}.

\newcommand{\hcell}[1]{\makecell[c]{\parbox{1.1cm}{\linespread{0.8}\selectfont\centering \textbf{#1}}}}
\begin{table*}[h]
\centering
\caption{Instance-level Pearson ($r$) and Kendall's $\tau$ correlation between model outputs and human annotations on Vectra-Test dataset. We compare our Vectra Model with: GLM-4.6V and GLM-4.6V-Flash \citep{GLM4.6V}; Qwen3-VL-235B-Instruct and Qwen3-VL-235B-Thinking \citep{Qwen3VL}; GPT-5, GPT-5-mini, and GPT-5-nano \citep{GPT5}; Claude-Sonnet-4 and Claude-Opus-4 \citep{Claude4}; Gemini-2.5-Flash, Gemini-2.5-Pro, Gemini 3 Flash, and Gemini-3-Pro \citep{Gemini2.5,Gemini3}.}
\label{tab:model_performance}
\renewcommand{\arraystretch}{0.9}
\setlength{\tabcolsep}{3pt}
\resizebox{\textwidth}{!}{
\begin{tabular}{lccccccccccccccc}
\toprule
\textbf{Model} & \hcell{Text\\Size} & \hcell{Text\\Color} & \hcell{Text\\position} & \hcell{Text Font\\Style} & \hcell{Text\\Layout} & \hcell{Text Pixel\\Clarity} & \hcell{Text\\Halluci\text{-}\\nation} & \hcell{Text\\Omission} & \hcell{Scene\\Size} & \hcell{Scene\\Color} & \hcell{Scene\\position}& \hcell{Scene Pixel\\Clarity} & \hcell{Scene\\Halluci\text{-}\\nation} & \hcell{Scene\\Omission} & \hcell{Vectra\\Score} \\
\midrule
\multicolumn{16}{c}{\textbf{Pearson ($r$)} $\uparrow$} \\
\midrule
GLM-4.6V & 0.0944 & 0.3473 & 0.3393 & 0.1893 & 0.2631 & 0.6141 & 0.6420 & 0.2951 & 0.2973 & 0.3909 & 0.3713 & 0.5373 & 0.5259 & 0.4817 & 0.7233 \\
GLM-4.6V-Flash & 0.2555 & 0.3854 & 0.3913 & 0.2681 & 0.2263 & 0.6021 & 0.3504 & 0.2258 & 0.3445 & 0.4473 & 0.4996 & 0.5750 & 0.4086 & 0.4900 & 0.7646 \\
Qwen3-VL-235B-Instruct & 0.0829 & 0.1988 & 0.2657 & 0.1944 & 0.1985 & 0.4673 & 0.5649 & 0.2008 & 0.1769 & 0.2621 & 0.2294 & 0.3114 & 0.2621 & 0.2638 & 0.5723 \\
Qwen3-VL-235B-Thinking & -0.0226 & 0.2429 & 0.2545 & 0.1018 & 0.2202 & 0.5569 & 0.5826 & 0.4307 & 0.3056 & 0.4015 & 0.3650 & 0.4925 & 0.4910 & 0.4253 & 0.6814 \\
GPT-5 & 0.1396 & 0.3909 & 0.3191 & 0.2396 & 0.2408 & 0.6990 & 0.5937 & 0.5895 & 0.3785 & 0.2816 & 0.3764 & 0.5948 & 0.4651 & 0.3173 & 0.7346 \\
GPT-5-mini & 0.2120 & 0.4828 & 0.3052 & 0.2852 & 0.2317 & 0.7337 & 0.5697 & 0.5401 & 0.3619 & 0.3311 & 0.3293 & 0.6301 & 0.4601 & 0.3337 & 0.6795 \\
GPT-5-nano & -0.0066 & 0.2584 & 0.2927 & 0.1061 & 0.1970 & 0.5308 & 0.4848 & 0.3879 & 0.3082 & 0.2445 & 0.3282 & 0.4657 & 0.3679 & 0.3258 & 0.5857 \\
Claude-Sonnet-4 & 0.1500 & 0.2589 & 0.4123 & 0.1912 & 0.1948 & 0.5861 & 0.5876 & 0.3391 & 0.1281 & 0.0989 & 0.1683 & 0.2255 & 0.0083 & 0.2051 & 0.5986 \\
Claude-Opus-4 & 0.1724 & 0.2100 & 0.4113 & 0.2128 & 0.2566 & 0.5307 & 0.5238 & 0.2934 & 0.1616 & 0.0775 & 0.2044 & 0.2165 & 0.0883 & 0.2339 & 0.5678 \\
Gemini-2.5-Flash & 0.0930 & 0.1974 & 0.3256 & 0.1338 & 0.2525 & 0.6506 & 0.5887 & 0.5420 & 0.2462 & 0.3318 & 0.3351 & 0.4090 & 0.3411 & 0.3099 & 0.6595 \\
Gemini-2.5-Pro & 0.1322 & 0.2667 & 0.3073 & 0.1790 & 0.2252 & 0.6935 & 0.6127 & 0.5464 & 0.4062 & 0.5563 & 0.4250 & 0.6326 & 0.6426 & 0.6029 & 0.7857 \\
Gemini-3-Flash & 0.2143 & 0.4293 & 0.3898 & 0.2395 & 0.2502 & 0.7207 & 0.6141 & 0.6654 & 0.4934 & 0.4656 & 0.5027 & 0.7133 & 0.7624 & 0.7271 & 0.8466 \\
Gemini-3-Pro & 0.1841 & 0.4275 & 0.3174 & 0.3171 & 0.2428 & 0.7292 & 0.6310 & 0.6374 & 0.4407 & 0.3446 & 0.4479 & 0.6276 & 0.6200 & 0.6667 & 0.8186 \\
\textbf{Vectra Model (4B)} & \textbf{0.4948} & \textbf{0.5124} & \textbf{0.6397} & \textbf{0.3487} & \textbf{0.2664} & \textbf{0.8006} & \textbf{0.8662} & \textbf{0.7260} & \textbf{0.6915} & \textbf{0.7346} & \textbf{0.7756} & \textbf{0.7204} & \textbf{0.8171} & \textbf{0.8126} & \textbf{0.8955} \\
\midrule
\multicolumn{16}{c}{\textbf{Kendall ($\tau$)} $\uparrow$} \\
\midrule
GLM-4.6V & 0.0880 & 0.3203 & 0.3382 & 0.1667 & \textbf{0.2541} & 0.6160 & 0.5968 & 0.2827 & 0.2748 & 0.3873 & 0.3563 & 0.5396 & 0.4959 & 0.4572 & 0.5673 \\
GLM-4.6V-Flash & 0.2280 & 0.3783 & 0.3771 & 0.2499 & 0.2265 & 0.5747 & 0.3319 & 0.2191 & 0.3252 & 0.4342 & 0.4679 & 0.5465 & 0.3795 & 0.4577 & 0.5787 \\
Qwen3-VL-235B-Instruct & 0.0661 & 0.1617 & 0.2417 & 0.1666 & 0.1819 & 0.4690 & 0.5268 & 0.1974 & 0.1677 & 0.2459 & 0.2226 & 0.3138 & 0.2543 & 0.2517 & 0.4665 \\
Qwen3-VL-235B-Thinking & -0.0161 & 0.2354 & 0.2619 & 0.0818 & 0.2051 & 0.5539 & 0.5462 & 0.4040 & 0.2976 & 0.3873 & 0.3570 & 0.4853 & 0.4693 & 0.4027 & 0.4992 \\
GPT-5 & 0.1311 & 0.3608 & 0.3195 & 0.2060 & 0.2327 & 0.6484 & 0.5534 & 0.5519 & 0.3429 & 0.2737 & 0.3476 & 0.5703 & 0.4463 & 0.3080 & 0.5432 \\
GPT-5-mini & 0.1910 & 0.4284 & 0.2794 & 0.2454 & 0.2248 & 0.6821 & 0.5291 & 0.5059 & 0.3379 & 0.3100 & 0.3105 & 0.6089 & 0.4132 & 0.3201 & 0.4926 \\
GPT-5-nano & -0.0005 & 0.2733 & 0.3017 & 0.0791 & 0.2043 & 0.5641 & 0.4391 & 0.3609 & 0.2842 & 0.2279 & 0.3210 & 0.4567 & 0.3293 & 0.3002 & 0.4026 \\
Claude-Sonnet-4 & 0.1322 & 0.2307 & 0.3726 & 0.1679 & 0.1927 & 0.5637 & 0.5509 & 0.3126 & 0.1197 & 0.0934 & 0.1549 & 0.2308 & 0.0147 & 0.1843 & 0.4399 \\
Claude-Opus-4 & 0.1671 & 0.2111 & 0.3757 & 0.1835 & 0.2403 & 0.5183 & 0.4867 & 0.2842 & 0.1532 & 0.0865 & 0.1944 & 0.2340 & 0.0728 & 0.2219 & 0.3756 \\
Gemini-2.5-Flash & 0.0779 & 0.1881 & 0.2880 & 0.1094 & 0.2333 & 0.6346 & 0.5640 & 0.5056 & 0.2483 & 0.3312 & 0.3045 & 0.4144 & 0.3215 & 0.2894 & 0.4712 \\
Gemini-2.5-Pro & 0.1316 & 0.2510 & 0.2925 & 0.1518 & 0.2153 & 0.6697 & 0.5854 & 0.5248 & 0.3741 & 0.5321 & 0.3873 & 0.6078 & 0.6079 & 0.5670 & 0.5844 \\
Gemini-3-Flash & 0.2049 & 0.4313 & 0.3913 & 0.2079 & 0.2309 & 0.6865 & 0.5786 & 0.6253 & 0.4518 & 0.4703 & 0.4836 & 0.6901 & 0.7254 & 0.6927 & 0.6440 \\
Gemini-3-Pro & 0.1872 & 0.4489 & 0.3438 & 0.3078 & 0.2136 & 0.6777 & 0.5939 & 0.6037 & 0.4006 & 0.3286 & 0.4196 & 0.6090 & 0.5962 & 0.6411 & 0.6155 \\
\textbf{Vectra Model (4B)} & \textbf{0.4803} & \textbf{0.4867} & \textbf{0.5812} & \textbf{0.3216} & 0.2424 & \textbf{0.7835} & \textbf{0.8137} & \textbf{0.6822} & \textbf{0.6655} & \textbf{0.7276} & \textbf{0.7446} & \textbf{0.7229} & \textbf{0.7835} & \textbf{0.7753} & \textbf{0.7243} \\
\bottomrule
\end{tabular}
}
\vspace{-0.4cm}
\end{table*}

\section{Experiments}
\label{sec:experiments}

To comprehensively validate Vectra, our experiments address four core research questions: \textbf{RQ1:} Does Vectra provide a more effective assessment of e-commerce image translation visual quality compared to existing evaluation schemes, both on in-domain and out-of-domain? \textbf{RQ2:} Does the Vectra Score design demonstrate reliability and effectively mitigate evaluation noise from subjective variability? \textbf{RQ3:} How does the Vectra Model, fine-tuned on our specialized data, perform against leading proprietary and open-source models? \textbf{RQ4:} What is the contribution of each component in our proposed method to the overall performance?

\begin{table}[H]
\scriptsize
\centering
\setlength{\tabcolsep}{16pt}
\renewcommand{\arraystretch}{0.9}
\caption{\textbf{Statistics of Vectra Datasets.} Stream A is fixed for benchmarking, while Stream B supports model development.}
\label{tab:dataset_stats}
\begin{tabular}{@{}lccc@{}}
\toprule
\textbf{Dataset Split} & \textbf{Count} & \textbf{Annotation} & \textbf{Usage} \\ \midrule
\multicolumn{4}{@{}l}{\textit{Stream A: Benchmark Evaluation}} \\
\textbf{Vectra-Bench} & 2,000 & Human Ranking & \textbf{RQ1} \\
\textbf{Vectra-Bench} & 200$^*$ & Human Scoring & \textbf{RQ2} \\ \midrule
\multicolumn{4}{@{}l}{\textit{Stream B: Model Evaluation (i.i.d.)}} \\
\textbf{Vectra-SFT} & 30,000 & Gemini-2.5-Pro & -- \\
\textbf{Vectra-RL} & 2,500 & Human Scoring & -- \\
\textbf{Vectra-Test} & 1,000 & Human Scoring & \textbf{RQ3, RQ4} \\ \bottomrule
\multicolumn{4}{@{}l}{$^*$Subset with distinct translation pairs for empirical study.}
\end{tabular}
\vspace{-0.5cm}
\end{table}

\subsection{Experimental Setup}
We construct two data streams from real-world product images (Table~\ref{tab:dataset_stats}): \textbf{Stream A} (2K images for Vectra-Bench) and \textbf{Stream B} (33.5K samples split into 30K SFT, 2.5K RL, and 1K test after distribution balancing). E-commerce experts with $>$2 years experience annotate in-domain data (RQ1-4), while crowdsourced non-expert annotators handle OOD evaluation (for RQ1). Rankings use pairwise comparisons with majority voting; scores use mode aggregation with ties broken by the minimum value. We address \textbf{RQ1} via \textbf{comparative evaluation}, \textbf{RQ2} via \textbf{empirical study}, \textbf{RQ3} via \textbf{model performance evaluation}, and \textbf{RQ4} via \textbf{ablation study}.

\subsection{Comparative Evaluation}
\label{sec:comparative_evaluation}
We evaluate whether Vectra can effectively assess IIMT quality by comparing its ranking ability against existing evaluation methods.

We conduct experiments on two benchmarks:
(1) \textbf{Vectra-Bench (In-Domain)}: 2,000 e-commerce images.
(2) \textbf{MCiT (Out-of-Domain)}: The MCiT \cite{MCiT} dataset covering documents, posters, and scene images to assess generalization.

We compare our Vectra method against both reference-based and reference-free metrics. For reference-based metrics, we employ SSIM, PSNR, LPIPS, and FID, which are widely used in IIMT or related works \cite{lantranslatotron,tian2025prim,tian2025exploring,IQA2,IQA1,IQA3}. Since the current IIMT evaluation typically requires reference images, we compute these metrics using the top-ranked translation of each sample (as determined by human experts) as the reference and calculating scores for the remaining candidates---a setup that provides reference-based methods for fair comparison. For reference-free metrics, we include SAMScore \cite{SAM-Score}, E-Commerce VQA \cite{ECO_VQA}, and HCIIT-Judge \cite{fu2024ensuring}. Specifically, we adapt E-Commerce VQA (originally designed for single-image quality assessment) to evaluate translations by computing quality score differences between translated and source images. HCIIT-Judge is a model-as-judge method specifically for IIMT evaluation.

To establish translation visual quality rankings, we employ 7 representative IIMT systems to generate diverse translation outputs across 5 language pairs (zh$\rightarrow$en/es/fr/ja/pt, 400 each): 3 proprietary e-commerce solutions (AIDC Image Translation Pro/Standard \cite{AIDC2025}, GhostCut \cite{GhostCut2025}) and 4 general-purpose multimodal models (GPT-Image-1 \cite{GPTImage1}, Nano Banana \cite{NanoBanana}, Nano Banana Pro \cite{NanoBananaPro}, Qwen-MT-Image \cite{QwenMTImage}). We then recruit 5 e-commerce experts for Vectra-Bench and 5 general annotators for MCiT to rank these 7 system outputs for each image through pairwise comparisons, with comparisons determined by a preference majority voting method.

Table~\ref{tab:sample_correlation} shows Vectra achieves the highest correlation on Vectra-Bench (Overall $r=0.738, \tau=0.637$), outperforming all baselines. On MCiT, Vectra maintains strong performance ($r=0.558, \tau=0.466$), particularly in Document and Poster scenarios. While SSIM performs competitively on Scene images (photographic content with minimal text), Vectra consistently captures quality across all content types, demonstrating robustness beyond e-commerce.

\subsection{Empirical Study}
We compare Vectra Score against HCIIT \cite{fu2024ensuring}, which provides detailed visual quality guidelines with a 3-point scale. Five experts annotated 200 samples under both protocols.

\begin{figure}[htbp]
\centering
\includegraphics[width=0.88\columnwidth]{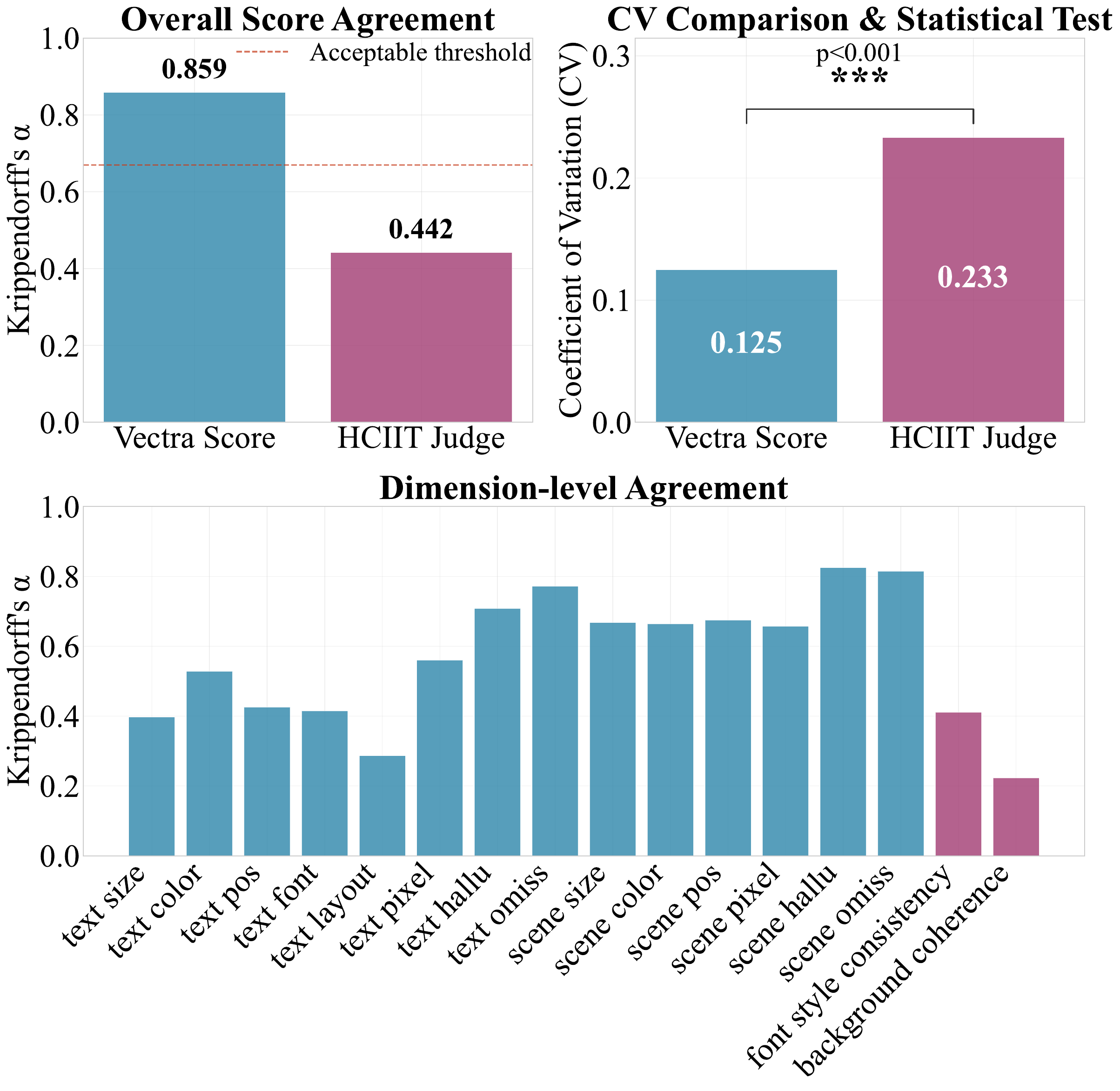}
\vspace{-0.3cm}
\caption{Top: overall inter-rater agreement (Krippendorff's $\alpha$) and Coefficient of Variation (CV) between Vectra Score and the baseline. Bottom: per-dimension $\alpha$.}
  \label{fig:empirical_study_results}
  \vspace{-0.6cm}
\end{figure}

As shown in Figure~\ref{fig:empirical_study_results}, Vectra achieves Krippendorff's $\alpha$ of 0.859 versus 0.442 for HCIIT, with 46\% CV reduction ($p < 0.001$). Dimension-level analysis reveals that Fidelity dimensions (Hallucination, Omission) drive consistency ($\alpha \in [0.70, 0.82]$), confirming that decomposing quality into observable defects reduces subjective ambiguity.

\subsection{Model Performance Evaluation}
We evaluate Vectra Model (4B parameters) against leading open-source and closed-source MLLMs on the Vectra-Test dataset, using the prompt described in Appendix~\ref{app:scoring_protocol}. We measure instance-level Pearson correlation ($r$) and Kendall's $\tau$ between model outputs and human annotations across 14 dimensions and the overall Vectra Score.

As shown in Table~\ref{tab:model_performance}, Vectra Model outperforms all compared models despite having significantly fewer parameters. Notably, Vectra Model achieves the best performance on all dimensions and the overall Vectra Score, with the only exception being a slightly lower Kendall's $\tau$ on Text Layout compared to GLM-4.6V. On the final Vectra Score, compared to the best-performing baseline Gemini 3 Flash, Vectra Model achieves 4.27\% higher Pearson correlation and 12.72\% higher Kendall's $\tau$. This demonstrates that our systematic data construction and model training pipeline enables it to achieve competitive evaluation capabilities. We further evaluate the truthfulness of model-generated reasoning in Appendix~\ref{app:reason_quality}, and provide detailed evaluation examples of Vectra Model in Appendix~\ref{app:assessment_examples}.

\subsection{Ablation Study}

We systematically ablate key components or process of Vectra Model, with each variant trained from scratch using identical hyperparameters. Results are shown in Table~\ref{tab:ablation}.
 
We investigate three major design choices: (1) \textit{Metric Design}: comparing uniform averaging against our multiplicative aggregation score; (2) \textit{Data Process}: evaluating the impact of our distribution balancing pipeline; (3) \textit{Model Training}: examining the contribution of preference alignment (RL) and structured reasoning capabilities.

\begin{table}[htbp]
\vspace{-0.1cm}
\centering
\scriptsize
\renewcommand{\arraystretch}{0.9}
\caption{Ablation study results on Vectra-Test dataset. We evaluate the contribution of each component or process by removing them individually from the full solution.}
\label{tab:ablation}
\setlength{\tabcolsep}{5pt}
\begin{tabular}{lcc}
\toprule
\textbf{Configuration} & \textbf{Pearson ($r$)} $\uparrow$ & \textbf{Kendall ($\tau$)} $\uparrow$ \\
\midrule
\multicolumn{3}{l}{\textit{Metric Design}} \\
uniform averaging over 14 dims & 0.8827 & 0.7153 \\
\midrule
\multicolumn{3}{l}{\textit{Data Process}} \\
w/o distribution balance & 0.8412 & 0.5949 \\
\midrule
\multicolumn{3}{l}{\textit{Model Training}} \\
w/o RL (SFT only) & 0.6739 & 0.4723 \\
\midrule
w/o reasoning & \textbf{0.9019} & \textbf{0.7348} \\
Vectra Model & \textbf{0.8955} & \textbf{0.7243} \\
\bottomrule
\end{tabular}
\vspace{-0.1cm}
\end{table}

The results reveal that distribution balancing and preference alignment are the most critical components, with their removal causing substantial performance degradation. Our multiplicative aggregation score provides modest but consistent improvement over uniform averaging. We also examine the effect of structured reasoning. The variant without reasoning achieves slightly higher correlation (Pearson $r$: 0.9019 vs. 0.8955), aligning with \cite{MCOT} that CoT-style reasoning can negatively affect perception-heavy multimodal tasks. In our framework, reasoning serves an indispensable role in quality assessment by providing interpretability-enabling error diagnosis and severity quantification.

\section{Related Work}
\subsection{In-Image Machine Translation Evaluation}
Text Image Translation (TIT) \cite{TIT-3,TIT-4,TIT-1,TIT-2,MCiT}, a subset of IIMT focusing on textual content, primarily evaluates translation accuracy using machine translation metrics such as BLEU \cite{BLEU}, METEOR \cite{METEOR}, and COMET \cite{COMET}. This text-centric paradigm has been inherited by early neural IIMT methods \cite{IIMT-1,tian2025exploring,lantranslatotron,tian2025prim}, which adopt visual quality metrics from image-to-image translation (I2I), such as SSIM \cite{SSIM} and FID\cite{FID}. However, these metrics capture only global similarity without diagnosing fine-grained visual defects. Recent work emphasizing font style and background quality has begun leveraging MLLMs as model-as-judge evaluators \cite{fu2024ensuring}. Meanwhile, IIMT has seen rapid industrial adoption, with LLM/MLLM-based systems \cite{GhostCut2025,AIDC2025} now powering cross-border e-commerce platforms globally. Yet neither approach provides fine-grained, interpretable diagnostics for scenarios with stringent visual quality requirements.

\subsection{Image-to-Image Translation Evaluation}
IIMT visual quality assessment inherently involves source-target image comparison, which crosses into image-to-image (I2I) translation evaluation \cite{I2I-1,I2I-2,IQA1,IQA3}. Widely used metrics from this domain include pixel-level measures such as PSNR and SSIM \cite{SSIM}, learned perceptual metrics like LPIPS \cite{LPIPS}, and distributional metrics such as FID \cite{FID}. Inspired by the success of foundation model-based metrics in other domains---BERTScore \cite{BertScore} for text similarity and CLIPScore \cite{ClipScore} for text-to-image alignment---recent I2I work has proposed SAMScore \cite{SAM-Score}, which leverages segmentation priors for structure-aware assessment. Domain-specific adaptations have also emerged, such as tailored metrics for medical imaging \cite{MR-I2I-eval}.

\section{Conclusion}
E-commerce IIMT presents unique challenges for visual quality evaluation: it demands multimodal defect analysis of textual and visual elements across paired product images with context-rich product information, where visual quality directly impacts commercial outcomes. Our experiments reveal that leading MLLMs struggle with reliable assessment in this task, creating a critical bottleneck for both production system benchmarking and preference-based model optimization. We present Vectra, an evaluation framework that addresses these limitations through a structured, multidimensional metric, which reduces the CV by 46\% compared to the previous method. Additionally, we introduce a domain-specialized 4B-parameter model that performs auto-evaluation with explainability on our benchmark. Through reasoning distillation, preference alignment, and systematic data curation on real-world e-commerce data, Vectra surpasses GPT-5 and Gemini-3 in human alignment scoring performance ($r=0.895$, $\tau=0.724$). Furthermore, Vectra outperforms mainstream visual quality assessment methods in ranking IIMT systems ($r=0.738$, $\tau=0.637$). Beyond e-commerce applications, out-of-distribution experiments demonstrate that Vectra generalizes well to broader IIMT scenarios. More broadly, our work reveals that perception-heavy cross-image tasks—a notable blind spot for many general-purpose MLLMs—benefit from specialized optimization, enabling efficient models to achieve superior preference alignment over their larger counterparts.

\section*{Impact Statement}
The quality of In-Image Machine Translation in e-commerce must align with both commercial requirements and human visual preferences. Among existing evaluation approaches, Vectra achieves the highest consistency with human perceptual assessments of e-commerce IIMT visual quality, demonstrating its potential for supervising and controlling the quality of commercial image translation systems. As cross-border e-commerce continues to expand globally, the need for reliable visual quality assessment becomes increasingly critical-not only for commercial efficiency but also for promoting equitable access to product information across linguistic boundaries. Vectra provides fine-grained diagnostic capabilities and serves as an effective reward signal for IIMT system optimization. By releasing our dataset and model publicly under a research-only license, we aim to accelerate research in this underexplored yet socially impactful domain. To ensure responsible data release, all images undergo automated PII detection; merchant logos and brand names are retained as they are essential for visual quality assessment but are not linked to merchant identities in our metadata. The dataset was collected from consenting merchants under data protection compliance, and all processing adheres to applicable privacy regulations. We will continue to advance this research direction, focusing on expanding language coverage and improving the framework's applicability to diverse e-commerce scenarios.


\bibliography{example_paper}
\bibliographystyle{icml2026}

%%%%%%%%%%%%%%%%%%%%%%%%%%%%%%%%%%%%%%%%%%%%%%%%%%%%%%%%%%%%%%%%%%%%%%%%%%%%%%%
%%%%%%%%%%%%%%%%%%%%%%%%%%%%%%%%%%%%%%%%%%%%%%%%%%%%%%%%%%%%%%%%%%%%%%%%%%%%%%%
% APPENDIX
%%%%%%%%%%%%%%%%%%%%%%%%%%%%%%%%%%%%%%%%%%%%%%%%%%%%%%%%%%%%%%%%%%%%%%%%%%%%%%%
%%%%%%%%%%%%%%%%%%%%%%%%%%%%%%%%%%%%%%%%%%%%%%%%%%%%%%%%%%%%%%%%%%%%%%%%%%%%%%%
\newpage
\appendix
\onecolumn

\section{Scoring Protocol}
\label{app:scoring_protocol}

\subsection{Multidimensional Quality Dimensions}
\label{app:dimensions}

The Vectra Score framework decomposes visual quality assessment into 14 fine-grained dimensions, organized along two axes: \textit{Content Type} (Textual vs. Scene) and \textit{Error Category} (Accuracy vs. Style). Table~\ref{tab:vectra_dimensions} presents the complete taxonomy.

\begin{table}[htbp]
\centering
\small
\caption{Taxonomy of evaluation dimensions in the Vectra Score framework. Dimensions are categorized by content type (Textual/Scene) and error category (Accuracy/Style).}
\label{tab:vectra_dimensions}
\setlength{\tabcolsep}{8pt}
\begin{tabular}{p{5.5cm} p{6cm} c}
\toprule
\textbf{Content Type} & \textbf{Dimension} & \textbf{Category} \\
\midrule
\multirow{8}{=}{{Textual Visual Quality}} 
  & Text Size Consistency & Style \\
  & Text Color Consistency & Style \\
  & Text Position Consistency & Style \\
  & Text Font Style Consistency & Style \\
  & Text Layout Consistency & Style \\
  & Text Pixel Clarity Consistency & Style \\
  & Text Hallucination & Accuracy \\
  & Text Omission & Accuracy \\
\midrule
\multirow{6}{=}{{Scene Visual Quality}} 
  & Scene Size Consistency & Style \\
  & Scene Color Consistency & Style \\
  & Scene Element Position & Style \\
  & Scene Pixel Clarity Consistency & Style \\
  & Scene Hallucination & Accuracy \\
  & Scene Omission & Accuracy \\
\bottomrule
\end{tabular}
\end{table}

For each dimension, the annotators assess the Defect Area Ratio (DAR) and assign a score $s \in \{1, 2, 3\}$ following the quantization scheme defined in \S\ref{sec:dar}. The \textit{Accuracy} dimensions (Hallucination, Omission) capture content fidelity errors, while \textit{Style} dimensions assess visual presentation quality.

\subsection{Annotation Guidelines and Prompts}
\label{app:annotation}

To evaluate the visual quality of e-commerce in-image machine translation, we employ a hybrid approach utilizing both MLLM and human annotations. To ensure consistency across these two modalities, we adhere to a unified annotation guideline.

\paragraph{Human Annotation Protocol.}
For human annotation, we recruit e-commerce experts with $>$2 years of domain experience. Each sample is assessed by five independent annotators to mitigate subjective variance. The final score is determined by the statistical mode of these ratings. In cases where multiple modes exist (i.e., a tie in frequency), we select the lowest score to enforce a conservative quality assessment. Notably, annotators rely on visual estimation rather than pixel-level computation when assessing DAR, follow the guideline in Table~\ref{tab:annotation_rubric}

\paragraph{MLLM Scoring Prompt.}
For automated annotation using Gemini-2.5-Pro, we design a structured prompt that explicitly incorporates the dimension definitions from Table~\ref{tab:vectra_dimensions} and the DAR-based scoring rubrics. The complete prompt is presented in Table~\ref{tab:gemini_prompt}.

% ... Table for gemini_prompt goes here ...

\begin{table*}[htbp]
\centering
\small
\caption{Human annotation guideline for evaluating the visual quality in our work. Annotators assess translation pairs across 14 dimensions using a 3-point scale based on visual defect severity.}
\begin{tabular}{|p{15.5cm}|}
\hline
\textbf{Human Annotation Guideline} \\
\textbf{Task}: Help us assess the visual quality of our machine-translated e-commerce images. Please carefully compare the generated image (right) against the source (left), focusing on the rendering quality of both textual and non-textual regions. \\[0.5em]

\textbf{Note}: This is purely a visual assessment task. You do not need foreign language proficiency, as you will focus on visual rendering quality rather than semantic translation accuracy. \\[0.5em]

\textbf{Scoring System} \\
You will use a 3-point scale based on the Defect Area Ratio (DAR), which measures the percentage of the relevant region that contains visible defects. Please note that you do not need to measure the area. DAR serves as a reference method for your visual estimation, helping you choose the appropriate score and minimizing subjective inconsistency. \\[0.5em]

\textbf{Understanding regions}: \\
\quad Text region: All areas containing text in the image \\
\quad Scene region: All non-text areas, including background, products, and design elements \\[0.5em]
\textbf{How to calculate DAR}: \\
\quad For text metrics: DAR is the proportion of defective text area compared to the total text area \\
\quad For scene metrics: DAR is the proportion of defective scene area compared to the total non-text area \\[0.5em]
\textbf{Apply the following scores}: \\
\quad 3-Excellent: $\text{DAR} \approx 0\%$ (No visible errors) \\
\quad 2-Fair: $0\% < \text{DAR} \le 30\%$ (Minor issues) \\
\quad 1-Poor: $\text{DAR} > 30\%$ (Significant issues) \\[0.5em]
% \textbf{Special rule}: If the source image contains \textbf{no text}, automatically assign \textbf{score 3} to all text metrics and focus only on scene metrics. \\[0.5em]

\textbf{Evaluation Metrics (14 dimensions)} \\
Evaluate each translation independently across the following dimensions. \\

\quad Text Size Consistency (t\_size): Text visual size in Trans matches Source. \\
\quad Text Color Consistency (t\_color): Text font color in Trans matches Source. \\
\quad Text Position Consistency (t\_pos): Text visual location in Trans aligns with Source coordinates. \\
\quad Text Font Style Consistency (t\_font): Font style (artistic effects) in Trans retains the original Source style. \\
\quad Text Layout Consistency (t\_layout): Text arrangement (vertical/horizontal/curved) in Trans follows the Source layout. \\
\quad Text Pixel Clarity Consistency (t\_pixel): Text sharpness in Trans is comparable to Source quality. \\
\quad Text Hallucination (t\_hallu): Trans is free from hallucinations, including distortions (semantic deviation, text anomalies) that compromise Source meaning, or spurious text not in Source. \\
\quad Text Omission (t\_omiss): No text content existing in Source is missing or left untranslated in Trans. \\
\quad Scene Size Consistency (s\_size): The size and scale of the global image and specific elements in Trans match Source. \\
\quad Scene Color Consistency (s\_color): Scene color in Trans matches Source and is free from removal artifacts/smudges. \\
\quad Scene Element Position Consistency (s\_pos): The positions of products, patterns, and elements in Trans align with Source. \\
\quad Scene Pixel Clarity Consistency (s\_pixel): Scene visual resolution/clarity in Trans is consistent with Source. \\
\quad Scene Hallucination (s\_hallu): Trans is free from scene hallucinations, including fidelity-compromising distortions (warping, anomalies) or spurious artifacts not in Source. \\
\quad Scene Omission (s\_omiss): No visual elements or objects existing in Source are missing in Trans. \\[0.5em]

\textbf{How to Annotate} \\
For each of the 14 metrics, assign exactly one label: \texttt{3-Excellent} (no visible issues), \texttt{2-Fair} (minor issues $\le30\%$ of area), or \texttt{1-Poor} (significant issues $>30\%$ of area). Always use complete label format (number and word) for consistency. \\[0.5em]

\textbf{Common Questions} \\
Q: What if source has no text? A: Assign score 3 to all text metrics; focus on scene metrics. \\
Q: What if multiple problems exist? A: Evaluate each metric independently. One image can receive low scores across multiple metrics. \\
% \textbf{Q: How to distinguish hallucination, color, and clarity?} A: Focus on what you observe: \textbf{Hallucination} = distortions/deformations/fabricated elements; \textbf{Color} = mismatched colors vs. source; \textbf{Clarity} = differences in sharpness/resolution. Score each metric independently. \\
\hline
\end{tabular}
\label{tab:annotation_rubric}
\end{table*}

\begin{table*}[htbp]
\centering
\small
\caption{Prompt for automated visual quality evaluation via MLLMs in our work. Language codes are dynamically mapped: \texttt{zh}$\rightarrow$Chinese, \texttt{en}$\rightarrow$English, \texttt{es}$\rightarrow$Spanish, \texttt{fr}$\rightarrow$French, \texttt{pt}$\rightarrow$Portuguese, \texttt{ja}$\rightarrow$Japanese.}
\begin{tabular}{|p{15.5cm}|}
\hline
\textbf{System Role}: You are an expert Visual Quality Evaluator for e-commerce In-Image Machine Translation. Your task is to assess the visual quality of an original source image (Source Image) and a translated product image (Trans Image). \\[0.5em]
\textbf{Translation Direction}: \{zh\} to \{en/es/fr/pt/ja\} \\[0.5em]
\textbf{Score}: Score 14 dimensions on a 3-point scale based on the Defect Area Ratio (DAR). If Source has no text, text dimensions default to 3. \\
\quad $\bullet$ Text Dimensions: $\text{DAR} = \text{Defective Text Area} / \text{Total Text Area}$ \\
\quad $\bullet$ Scene Dimensions: $\text{DAR} = \text{Defective Scene Area} / \text{Total Non-Text Area}$ \\
\quad 3-Excellent: $\text{DAR} \approx 0\%$ (No visible errors) \\
\quad 2-Fair: $0\% < \text{DAR} \le 30\%$ (Minor issues) \\
\quad 1-Poor: $\text{DAR} > 30\%$ (Significant issues) \\[0.5em]
\textbf{Evaluation Dimensions} (14 dimensions covering textual content and scene (non-textual content)): \\
\quad Text Size Consistency (t\_size): Text visual size in Trans matches Source. \\
\quad Text Color Consistency (t\_color): Text font color in Trans matches Source. \\
\quad Text Position Consistency (t\_pos): Text visual location in Trans aligns with Source coordinates. \\
\quad Text Font Style Consistency (t\_font): Font style (artistic effects) in Trans retains the original Source style. \\
\quad Text Layout Consistency (t\_layout): Text arrangement (vertical/horizontal/curved) in Trans follows the Source layout. \\
\quad Text Pixel Clarity Consistency (t\_pixel): Text sharpness in Trans is comparable to Source quality. \\
\quad Text Hallucination (t\_hallu): Trans is free from hallucinations, including distortions (semantic deviation, text anomalies) that compromise Source meaning, or spurious text not in Source. \\
\quad Text Omission (t\_omiss): No text content existing in Source is missing or left untranslated in Trans. \\
\quad Scene Size Consistency (s\_size): The size and scale of the global image and specific elements in Trans match Source. \\
\quad Scene Color Consistency (s\_color): Scene color in Trans matches Source and is free from removal artifacts/smudges. \\
\quad Scene Element Position Consistency (s\_pos): The positions of products, patterns, and elements in Trans align with Source. \\
\quad Scene Pixel Clarity Consistency (s\_pixel): Scene visual resolution/clarity in Trans is consistent with Source. \\
\quad Scene Hallucination (s\_hallu): Trans is free from scene hallucinations, including fidelity-compromising distortions (warping, anomalies) or spurious artifacts not in Source. \\
\quad Scene Omission (s\_omiss): No visual elements or objects existing in Source are missing in Trans. \\[0.5em]
\textbf{Strict Output XML Format}: \\
\quad \texttt{<t\_size\_reason>Example: \{CONTENT\} of translated image is \{ISSUE\} in \{POSITION\} region compared to source, impacting \{EFFECT\}\% of the area (DAR approx \{EFFECT\}\%).</t\_size\_reason>} \\
\quad \texttt{<t\_size\_score>\{"1-Poor" | "2-Fair" | "3-Excellent"\}</t\_size\_score>} \\
\quad \texttt{<t\_color\_reason>...</t\_color\_reason> <t\_color\_score>...</t\_color\_score>} \\
\quad \texttt{<t\_pos\_reason>...</t\_pos\_reason> <t\_pos\_score>...</t\_pos\_score>} \\
\quad \texttt{<t\_font\_reason>...</t\_font\_reason> <t\_font\_score>...</t\_font\_score>} \\
\quad \texttt{<t\_layout\_reason>...</t\_layout\_reason> <t\_layout\_score>...</t\_layout\_score>} \\
\quad \texttt{<t\_pixel\_reason>...</t\_pixel\_reason> <t\_pixel\_score>...</t\_pixel\_score>} \\
\quad \texttt{<t\_hallu\_reason>...</t\_hallu\_reason> <t\_hallu\_score>...</t\_hallu\_score>} \\
\quad \texttt{<t\_omiss\_reason>...</t\_omiss\_reason> <t\_omiss\_score>...</t\_omiss\_score>} \\
\quad \texttt{<s\_size\_reason>...</s\_size\_reason> <s\_size\_score>...</s\_size\_score>} \\
\quad \texttt{<s\_color\_reason>...</s\_color\_reason> <s\_color\_score>...</s\_color\_score>} \\
\quad \texttt{<s\_pos\_reason>...</s\_pos\_reason> <s\_pos\_score>...</s\_pos\_score>} \\
\quad \texttt{<s\_pixel\_reason>...</s\_pixel\_reason> <s\_pixel\_score>...</s\_pixel\_score>} \\
\quad \texttt{<s\_hallu\_reason>...</s\_hallu\_reason> <s\_hallu\_score>...</s\_hallu\_score>} \\
\quad \texttt{<s\_omiss\_reason>...</s\_omiss\_reason> <s\_omiss\_score>...</s\_omiss\_score>} \\
\hline
\end{tabular}
\label{tab:gemini_prompt}
\end{table*}

\section{Translation Image Construction}
\label{app:translation_prompt}
%我们介绍了Vectra Model一个领域对齐的模型裁判，它的训练依赖于Vectra SFT/RL/Meta Evaluation。这些数据的构建需要依赖电商产品图片构建翻译对，我们employ了AIDC Image Translation和AIDC Image Translation Pro两个电商领域的state of the art方法，这两个方法直接进行api指定翻译语言的调用就可以直接完成翻译图片的构造。另外我们还employ了一些通用MLLM Nano Banana Pro Nano Banana 和GPT-Image-1，这些模型应用了提示词tab:translation_prompt完成翻译图片的构造。
We present Vectra Model, a domain-aligned model judge whose training relies on the Vectra SFT, RL, and Meta Evaluation datasets. The construction of these datasets necessitates the generation of translation pairs derived from e-commerce product images. To this end, we leverage two state-of-the-art (SOTA) methods in the e-commerce domain: AIDC Image Translation and AIDC Image Translation Pro. These systems facilitate the direct generation of translated images via API calls by simply specifying the target language. Furthermore, we employ general-purpose MLLMs, specifically Nano Banana Pro, Nano Banana, and GPT-Image-1, to generate translation outputs guided by the prompts detailed in Table~\ref{tab:translation_prompt}.

\begin{table}[htbp]
\centering
\small
\caption{Translation prompt used for constructing IIMT samples from real-world e-commerce product images. The translation directions cover Chinese to English, Spanish, French, Portuguese, and Japanese.}
\begin{tabular}{|p{12cm}|}
\hline
Translate the text in the following image from \{source\_language\} to \{target\_language\} and replace the original text with the translated text in the image. \\
\hline
\end{tabular}
\label{tab:translation_prompt}

\end{table}

\section{Data Sampling Details}
\label{app:data_sampling}
%电商场景图片几乎都是套图即同一个商品或者排版风格有很多类似的图片，为了保证原始数据质量，我们采用了Segment Anything Model (SAM) - ViT Huge (ViT-H) version，它已在包含 1100 万张图像和 110 亿个掩码的数据集上进行训练，并在各种分割任务中表现出强大的零样本性能。具体而言，我们使用开源实现对输入图像进行批量预处理与前向编码。从视觉编码器的最后一层提取密集特征 $\mathbf{F}_i \in \mathbb{R}^{C\times H\times W}$，并通过全局平均池化得到紧凑的全局表征 $\mathbf{z}_i \in \mathbb{R}^{C}$：
% \begin{equation}
% \mathbf{z}_i=\frac{1}{HW}\sum_{h=1}^{H}\sum_{w=1}^{W}\mathbf{F}_i[:,h,w].
% \end{equation}
% 在 SAM ViT-H 配置下，$C\approx256$，该描述子在大规模电商图像上展现出稳定的语义聚合能力。为消除量纲与尺度差异，我们对所有 $\mathbf{z}_i$ 进行标准化：
% \begin{equation}
% \tilde{\mathbf{z}}_i=\frac{\mathbf{z}_i-\boldsymbol{\mu}}{\boldsymbol{\sigma}},\quad
% \boldsymbol{\mu}=\frac{1}{N}\sum_{i=1}^{N}\mathbf{z}_i,\quad
% \boldsymbol{\sigma}^2=\frac{1}{N}\sum_{i=1}^{N}(\mathbf{z}_i-\boldsymbol{\mu})^2,
% \end{equation}
% 实现上对应 StandardScaler 的拟合与变换。为保证可扩展性与稳健性，我们采用高效的批式推理与结果持久化策略，以在大规模数据上维持一致的特征质量与复现性。

% \subsection{Clustering and Representative Selection}
% 我们在标准化特征上执行 K-Means 聚类（簇数为 $K$），获得簇标签与中心 $\{\mathbf{c}_k\}_{k=1}^{K}$。对于每个簇，依据到簇中心的欧氏距离选取单一代表样本：
% \begin{equation}
% i^\ast=\arg\min_{i}\ \lVert \tilde{\mathbf{z}}_i-\mathbf{c}_k\rVert_2,
% \end{equation}
% 即保留距离中心最近的图像。该中心驱动的选择在同款或同版式的近重复簇内有效压缩冗余，同时在全局层面维持多模态分布的覆盖。

E-commerce imagery is predominantly characterized by product series, where multiple images of the same item or similar layout styles result in significant redundancy. To ensure data quality, we leverage the Segment Anything Model (SAM) with a ViT-H backbone, trained on SA-1B (11M images and 1.1B masks), which exhibits strong zero-shot performance across diverse segmentation tasks.
Specifically, we employ an open-source implementation to perform batch preprocessing and forward encoding. From the final layer of the visual encoder, we extract dense features $\mathbf{F}_i \in \mathbb{R}^{C \times H \times W}$ and apply global average pooling to obtain a compact global representation. Let $\mathbf{F}_i[c,h,w]$ denote the feature value at channel $c$ and spatial location $(h,w)$ for image $i$; the pooled descriptor is
\begin{equation} 
\mathbf{z}_i = \frac{1}{HW}\sum_{h=1}^{H}\sum_{w=1}^{W}\mathbf{F}_i[:,h,w] \in \mathbb{R}^{C}.
\end{equation}

Under the SAM ViT-H configuration used in our implementation, $C=256$, and this descriptor exhibits stable semantic aggregation on large-scale e-commerce images. To remove dimensional and scale discrepancies, we standardize all descriptors via
\begin{equation}
\tilde{\mathbf{z}}_i=\frac{\mathbf{z}_i-\mathbb{E}[\mathbf{z}]}{\sqrt{\operatorname{Var}[\mathbf{z}]}}\!,
\end{equation}
which is equivalent to standard score normalization.
We then perform K-means clustering on the normalized features to obtain cluster labels and centroids $\{\mathbf{c}_k\}_{k=1}^{K}$. For each cluster, we select a single representative image as the one closest to the centroid in Euclidean distance:
\begin{equation} 
i^\ast=\arg\min_{i\in\mathcal{C}_k}\ \lVert \tilde{\mathbf{z}}_i-\mathbf{c}_k\rVert_2.
\end{equation}

This centroid-driven selection effectively compresses redundancy within near-duplicate clusters (e.g., identical products or layouts) while maintaining coverage of the global multimodal distribution.

%\afterpage{
\begin{table*}[htbp]
\centering
\small
\caption{The exact structure and content of the prompt generation pipeline. Part A reflects the final string format presented to the Nano Banana Pro, where placeholders are dynamically populated by the programmatic logic detailed in Part B.}
\begin{tabular}{|p{15.5cm}|}
\hline
\textbf{Part A: Prompt Template} \\[0.5em]

\textbf{System Role}: \\
\quad You are an expert Image Fault Injector and Negative Sample Generator for e-commerce visuals. Your task is to take a valid Image, and intentionally modify the Image to introduce a specific defect based on the parameters below. \\[0.5em]

\textbf{Target Defect Introduction}: \\
\quad \{\texttt{defect\_detail}\} \\[0.5em]

\textbf{Definition of Defect Area Ratio (DAR)}: \\
\quad The Score is strictly defined by the coverage area of the error: \\
\quad \{\texttt{dar\_definition}\} \\[0.5em]

\{\texttt{dar\_constraint}\} \\[0.5em]

\textbf{Output}: \\
\quad Please provide the modified image, including the DAR-constrained significant target defect. \\
\hline
\end{tabular}

\begin{tabular}{|p{15.5cm}|}
\hline
\textbf{Part B: Programmatic Values for Placeholders} \\[0.5em]

\textbf{1. Mapping for} \{\texttt{defect\_detail}\} (\textit{Indicator Map}): \\
\quad $\bullet$ \textbf{t\_size}: Text Size Defect: Render text with drastically inconsistent scaling. Make some characters microscopically unreadable while others are comically huge, ignoring design balance. \\
\quad $\bullet$ \textbf{t\_color}: Text Color Defect: Force text colors to blend into the background (low contrast) or use neon/jarring colors that completely mismatch the original palette, making it hard to read. \\
\quad $\bullet$ \textbf{t\_pos}: Text Position Defect: Place text blocks in nonsensical areas. Force text to overlap on top of main subject faces, extend halfway off the canvas edge, or cover critical product details. \\
\quad $\bullet$ \textbf{t\_font}: Text Font Style Defect: Swap professional fonts with completely inappropriate styles (e.g., Comic Sans, messy handwriting, or jagged fonts) that destroy the visual consistency. \\
\quad $\bullet$ \textbf{t\_layout}: Text Layout Defect: Break the internal flow of text. Use chaotic line spacing (leading) where lines crash into each other, erratic character spacing (kerning), or incorrect reading direction (e.g., upside-down letters). \\
\quad $\bullet$ \textbf{t\_pixel}: Text Pixel Clarity Defect: Text Pixel Clarity Defect: Heavily pixelate or blur the text specifically. Simulate severe JPEG compression artifacts making the edges jagged and the characters fuzzy/illegible. \\
\quad $\bullet$ \textbf{t\_hallu}: Text Hallucination Defect: Insert random, nonsensical character strings, floating symbols, or repeated words where they do not belong, cluttering the image. \\
\quad $\bullet$ \textbf{t\_omiss}: Text Omission Defect: Randomly erase key words or sentences, leaving unnatural blank gaps or cut-off sentences in the middle of a paragraph. \\
\quad $\bullet$ \textbf{s\_size}: Scene Size Defect: Warp the aspect ratio of objects. Stretch products or people unnaturally wide or compress them to be thin, defying physics and perspective. \\
\quad $\bullet$ \textbf{s\_color}: Scene Color Defect: Apply unnatural color filters (e.g., green tint on skin, inverted colors on products) or wash out colors to look like a damaged image file. \\
\quad $\bullet$ \textbf{s\_pos}: Scene Element Position Defect: Break the logical relationship between objects. Place furniture on the ceiling, cars indoors, or detach a person from their shadow. Misalign interacting objects (e.g., a hand reaching for a cup but missing it by a foot). \\
\quad $\bullet$ \textbf{s\_pixel}: Scene Pixel Clarity Defect: Apply heavy noise, grain, or blocky compression artifacts to the background and objects, making the scene look like a low-quality thumbnail. \\
\quad $\bullet$ \textbf{s\_hallu}: Scene Hallucination Defect: Generate spurious entities that do not exist in reality. Create extra limbs on people, objects melting into each other, or random geometric shapes floating in the scene. \\
\quad $\bullet$ \textbf{s\_omiss}: Scene Omission Defect: Remove essential parts of objects (e.g., a car missing a wheel, a person missing a hand) leaving a confusing or empty space. \\[0.5em]

\textbf{2. Logic for} \{\texttt{dar\_definition}\}: \\
\quad $\bullet$ \textbf{If \texttt{Text Indicators (t\_*)}}: \texttt{DAR = Area of Defective Text / Total Text Area} \\
\quad $\bullet$ \textbf{If \texttt{Scene Indicators (s\_*)}}: \texttt{DAR = Area of Defective Scene / Total Non-Text Area} \\[0.5em]

\textbf{3. Logic for} \{\texttt{dar\_constraint}\} (\textit{Score-based strings}): \\
\quad $\bullet$ \textbf{Score 2}: \textbf{Strict DAR Constraint (Minor Error)}: You must ensure the Defect Area Ratio (DAR) is greater than 0\% but less than or equal to 30\% (0\% $<$ DAR $\le$ 30\%). \\
\quad $\bullet$ \textbf{Score 1}: \textbf{Strict DAR Constraint (Severe Error)}: You must ensure the Defect Area Ratio (DAR) is greater than 30\% (DAR $>$ 30\%). \\
\hline
\end{tabular}
\label{tab:gen_prompt}

\end{table*}
%}

%\afterpage{
\begin{figure*}[htbp]
  \centering
  \includegraphics[width=0.78\textwidth]{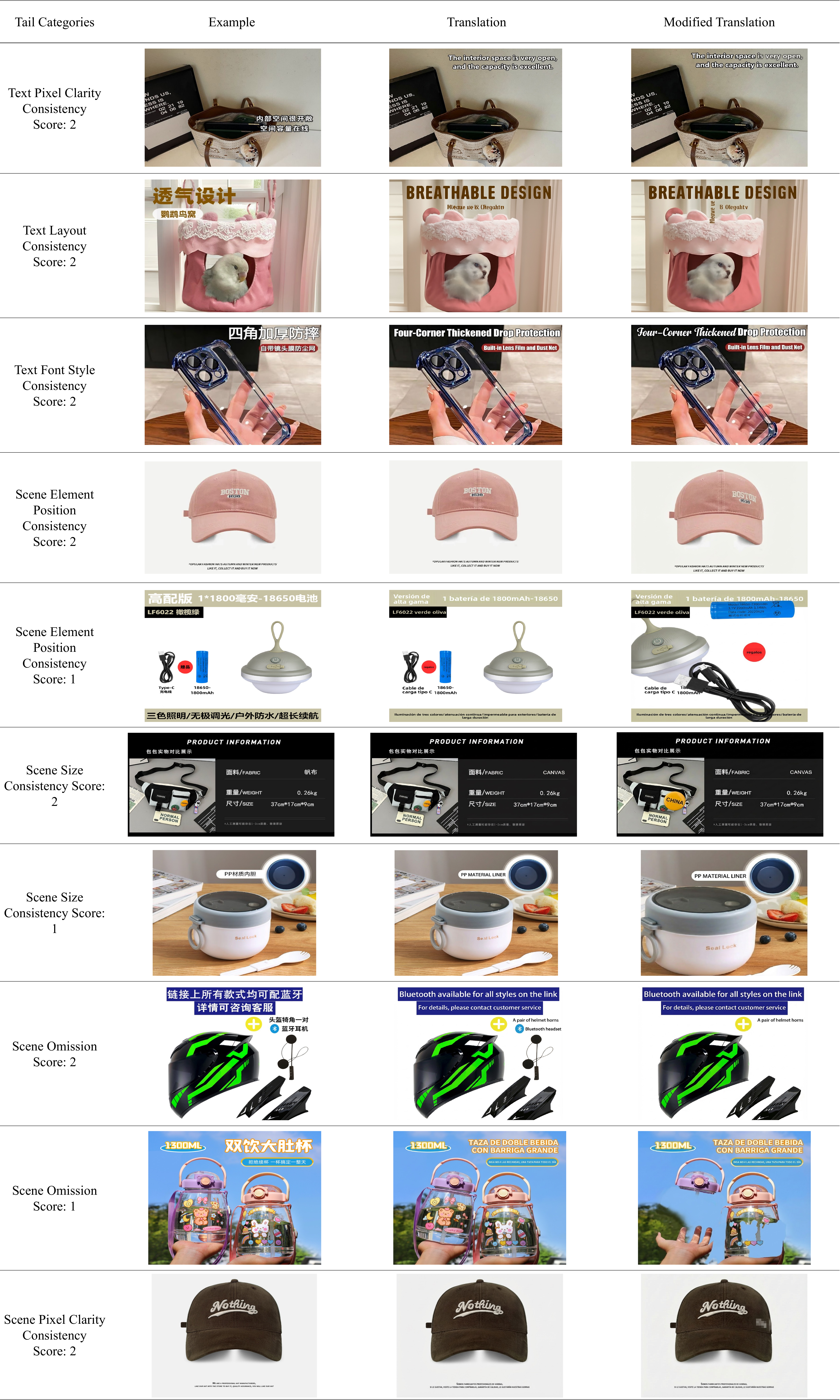}
  \caption{Qualitative examples generated based on the prompts described in Table \ref{tab:gen_prompt}. The figure illustrates the synthesis of Score 1 and Score 2 samples for long-tail categories (below $Q_1$) by injecting defects into Score 3 baselines.}
  \label{fig:gen_examples}
  
\end{figure*}
%}

\section{Distribution Balance Details}
\label{app:class_balance}
%在类均衡飞轮中，我们针对长尾的分值（稀缺）的数据采用Nano Banana Pro进行稀缺分值样本的重构，通过观察数据，没有问题的（3分）样本普遍极多，所有稀缺的分值都来源于1分或者2分，因此，为了避免大幅度的影响其它指标的原有评分引入更大的不确定性，最终选择直接抽取3分也就是没有任何问题的翻译样本去有针对性的构造有特定问题和程度的样本，构造prompt见tab:gen_prompt
In the Minority Augmentation, we utilize Nano Banana Pro to reconstruct samples for the long-tail (scarce) score categories. Analysis of the dataset distribution reveals a predominance of flawless samples (Score 3), with scarcity concentrated in the Score 1 and Score 2 categories. To generate negative samples without introducing uncertainty to other metric dimensions, we strictly sample from the pool of Score 3 (flawless) translations. These samples serve as baselines into which we inject specific defects of varying severity to synthesize Score 1 and Score 2 instances. The specific prompts governing this construction are provided in Table~\ref{tab:gen_prompt}.
%table XXXXXX

%在文中我们针对部分样本量小于下四分位点的分值样本进行构造的效果示例如图：
Figure~\ref{fig:gen_examples} presents qualitative examples derived from our data construction pipeline. We specifically visualize modified examples covering the tail categories with a sample count falling below the first quartile ($Q_1$) of the overall distribution, which were generated utilizing the prompting strategies detailed above.

%\section{Training Plots}
%\label{app:training_plots}
%\section{Dataset Statistics}
%\label{app:dataset_stats}
%统计最终经过类均衡和分层采样之后每个数据集到底有多少数据，可以的话可视化一下各个语言和各个指标score的分布

%Table~\ref{tab:dataset_stats} provides detailed statistics for each subset of the Vectra data suite.

\section{Training Configuration}
\label{app:training_config}

\subsection{Model Architecture}

Vectra model is initialized from Qwen3-VL-4B-Instruct \cite{Qwen3VL}, a 4B-parameter vision-language model. We perform full-parameter fine-tuning without any parameter freezing or adapter modules. The model processes paired images (source and translated) as concatenated visual inputs with variable resolutions preserved from the original images, along with task-specific textual prompts.

\subsection{Supervised Fine-Tuning}
\label{app:sft_config}
We first train Vectra Model on the Vectra-SFT dataset using standard cross-entropy loss. Table~\ref{tab:sft_config} summarizes the SFT configuration.

\begin{table}[htbp]
\centering
\small
\caption{Supervised fine-tuning hyperparameters.}
\label{tab:sft_config}
\begin{tabular}{@{}ll@{}}
\toprule
\textbf{Hyperparameter} & \textbf{Value} \\
\midrule
Base model & Qwen3-VL-4B-Instruct \\
Training epochs & 3 \\
Batch size (per device) & 1 \\
Gradient accumulation steps & 8 \\
Effective batch size & 64 \\
Learning rate & $2 \times 10^{-5}$ \\
Learning rate schedule & Cosine with warmup \\
Warmup ratio & 3\% of total steps \\
Optimizer & AdamW ($\beta_1$=0.9, $\beta_2$=0.999) \\
Weight decay & 0 \\
Gradient clipping & 1.0 \\
Max sequence length & 4096 \\
Image resolution & Native (variable) \\
Precision & bfloat16 \\
\bottomrule
\end{tabular}
\end{table}

\subsection{Preference Optimization}
\label{app:rl_config}
Following SFT, we apply GSPO \cite{GSPO} to align the model with human preferences. Table~\ref{tab:gspo_config} summarizes the hyperparameters. Our reward function combines structural compliance with human preference alignment:
\begin{equation}
r(y, y^*) = r_{\text{format}}(y) + r_{\text{preference}}(y, y^*),
\end{equation}
where $y$ is the model output and $y^*$ is the expert annotation.

\paragraph{Format Reward.}
To ensure parseable outputs, $r_{\text{format}}(y)$ evaluates structural compliance of the 14-dimension XML schema. For each dimension $d \in \{1, \ldots, 14\}$, we check: (1) presence of all four tags ($\langle d\_\text{reason}\rangle$, $\langle/d\_\text{reason}\rangle$, $\langle d\_\text{score}\rangle$, $\langle/d\_\text{score}\rangle$), awarding $+0.5$ per tag; (2) correct ordering where reason precedes score without intervening tags, awarding $+1.0$ per valid pair. The maximum format reward is $14 \times (4 \times 0.5 + 1.0) = 42$.

\paragraph{Preference Reward.}
Given expert-annotated ground-truth scores $\{s_d^*\}_{d=1}^{14}$ where $s_d^* \in \{1, 2, 3\}$, we extract predicted scores $\{\hat{s}_d\}$ from the model output $y$ and compute:
\begin{equation}
r_{\text{preference}}(y, y^*) = \sum_{d=1}^{14} \max(0, 3 - |s_d^* - \hat{s}_d|)
\end{equation}
This linear penalty awards 3 points for exact match, 2 for distance-1 error, 1 for distance-2, and 0 otherwise. The maximum preference reward is $14 \times 3 = 42$.

\begin{table}[htbp]
\centering
\small
\caption{GSPO preference optimization hyperparameters.}
\label{tab:gspo_config}
\begin{tabular}{@{}ll@{}}
\toprule
\textbf{Hyperparameter} & \textbf{Value} \\
\midrule
Base model & Vectra-SFT checkpoint \\
Training epochs & 3 \\
Batch size (per device) & 2 \\
Gradient accumulation steps & 2 \\
Effective batch size & 32 \\
Group size $G$ & 8 \\
Learning rate & $2 \times 10^{-6}$ \\
Learning rate schedule & Constant \\
KL penalty coefficient $\beta$ & 0.0 \\
Clipping threshold $\epsilon$ & $3 \times 10^{-4}$ \\
Upper clipping $\epsilon_{\text{high}}$ & $4 \times 10^{-4}$ \\
Sampling temperature & 1.3 \\
Max completion length & 2,048 \\
Optimizer & AdamW ($\beta_1$=0.9, $\beta_2$=0.999)\\
Precision & bfloat16 \\
\bottomrule
\end{tabular}
\end{table}

\subsection{Implementation Details}

\paragraph{Distributed Training.} All experiments utilize 8 NVIDIA A100 GPUs (80GB) with DeepSpeed ZeRO-3 \cite{deepspeed} for optimizer and parameter sharding. We enable CPU offloading and activation checkpointing to handle long sequence lengths and high-resolution images.

\paragraph{Generation Infrastructure.} During GSPO, we employ vLLM \cite{vllm} in colocated mode for efficient batched generation, with GPU memory utilization set to 0.25 to accommodate the generation of multiple responses per prompt.

\paragraph{Training Framework.} We implement GSPO using the Hugging Face TRL (Transformer Reinforcement Learning) library \cite{TRL}, configured specifically for the GSPO regime with zero KL regularization and sequence-level importance sampling. This configuration follows the recommendations in the GSPO paper for achieving better alignment without explicit KL constraints.

\paragraph{Numerical Stability.} We use bfloat16 precision throughout training for numerical stability and memory efficiency. Gradient clipping is applied with a threshold of 1.0 during SFT to prevent gradient explosion.

%%%%%%%%%%%%%%%%%%%%%%%%%%%%%%%%%%%%%%%%%%%%%%%%%%%%%%%%%%%%%%%%%%%%%%%%%%%%%%%
%%%%%%%%%%%%%%%%%%%%%%%%%%%%%%%%%%%%%%%%%%%%%%%%%%%%%%%%%%%%%%%%%%%%%%%%%%%%%%%

\section{Reasoning Quality Analysis}
\label{app:reason_quality}
We evaluate whether model-generated reasoning chains faithfully describe the multimodal defects when comparing source and translated image pairs. Specifically, we assess the truthfulness accuracy of the reasoning content---whether the described defects, their locations, and severity actually exist in the images. We sample 100 instances from Vectra-Test and recruit the expert group from Section~\ref{sec:experiments}, with each sample evaluated by at least one expert. Annotators judge whether all factual descriptions in the reasoning match the actual visual situation (binary: true/false).

\begin{figure}[htbp]
  \centering
  \includegraphics[width=0.8\columnwidth]{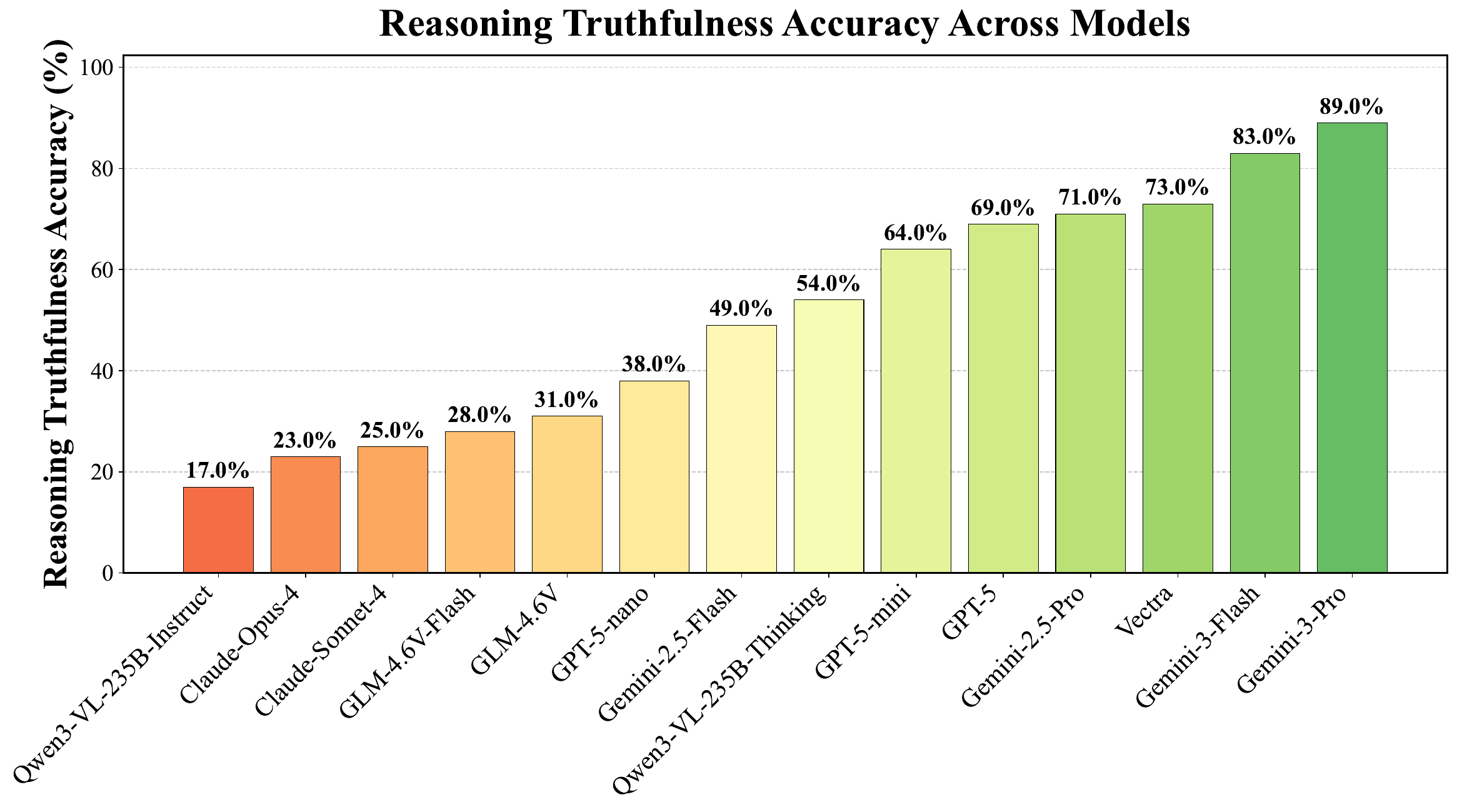}
  \caption{Reasoning truthfulness accuracy (\%) across models on 100 sampled instances.}
  \label{fig:reasoning_truthfulness}
\end{figure}

As shown in Figure~\ref{fig:reasoning_truthfulness}, analyzing multimodal defects across image pairs in context-dense e-commerce scenarios remains challenging for current MLLMs---8 out of 14 models achieve below 60\% accuracy. This underscores that perception-intensive cross-image comparison constitutes a notable blind spot for general-purpose MLLMs, where specialized optimization proves essential for reliable visual reasoning.

Vectra Model, trained on Gemini-2.5-Pro reasoning data and aligned via GSPO, achieves 73\% truthfulness accuracy, ranking third behind Gemini-3-Pro (89\%) and Gemini-3-Flash (83\%). Notably, Vectra outperforms its teacher model Gemini-2.5-Pro (71\%), demonstrating that preference alignment not only improves scoring calibration but also enhances reasoning quality. Combined with its superior scoring correlation (Table~\ref{tab:model_performance}), Vectra offers a favorable trade-off between reasoning truthfulness and assessment accuracy for practical deployment.

\section{Assessment Examples}
\label{app:assessment_examples}

We present representative evaluation samples from Vectra on both in-domain (e-commerce) and out-of-domain (document, poster, scene) scenarios to discuss the model's visual quality assessment capabilities.

\subsection{E-Commerce Scenarios}
%我们展示了一些有代表性的案例，涉及到了电商领域常见的auto IIMT系统容易造成的几种典型的缺陷以及Vectra的可解释评估结果， as illustrated in Image \cite{eco_example_1,eco_example_2,eco_example_3}。
%在\cite{eco_example_1}中，这种图片被一个传统的基于OCR图片文本识别，文字消除，背景填补，文字翻译，文字重新放入的pipline系统 AIB Image Translation，这类系统非常容易在文字消除过程中引入背景填补的缺陷，以及文字翻译后的颜色、字体、大小等部分出现问题，同时会影响图片原本的背景细节。根据我们的结果看出Vectra，在训练中已经掌握了一定的缺陷判别思维能力和可解释能力。
We present representative cases illustrating several typical defects commonly introduced by automated IIMT systems in e-commerce scenarios, along with Vectra's interpretable evaluation results.

\begin{figure*}[htbp]
  \centering
  \vspace{-0.3cm}
  \includegraphics[width=\textwidth]{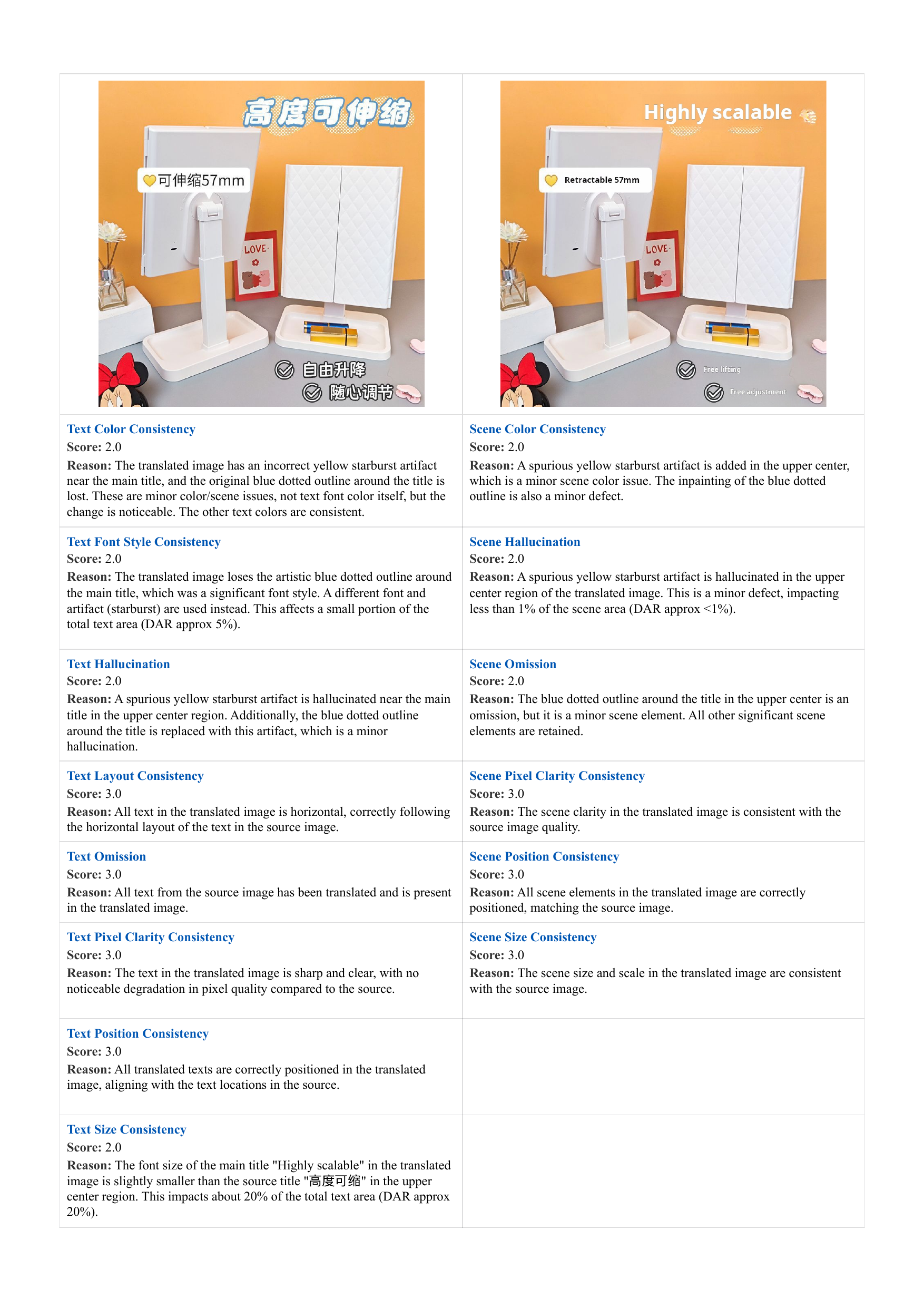}
  \vspace{-1.8cm}
  \caption{E-commerce evaluation example illustrating background inpainting artifacts and text rendering inconsistencies. Left: source image; Right: translated image.}
  \label{fig:eco_example_1}
\end{figure*}

Figure~\ref{fig:eco_example_1} demonstrates a product image processed by a traditional pipeline-based system that sequentially performs OCR text recognition, text erasure, background inpainting, text translation, and text re-rendering. Such pipeline systems are particularly prone to introducing background inpainting artifacts during the text erasure stage. Furthermore, the re-rendered text often exhibits inconsistencies in color, font style, and size compared to the original, which in turn compromises the surrounding background details of the image. As evidenced by our evaluation results, Vectra enables it to accurately identify and explain these cascading visual quality issues.

\begin{figure*}[htbp]
  \centering
  \vspace{-0.3cm}
  \includegraphics[width=\textwidth]{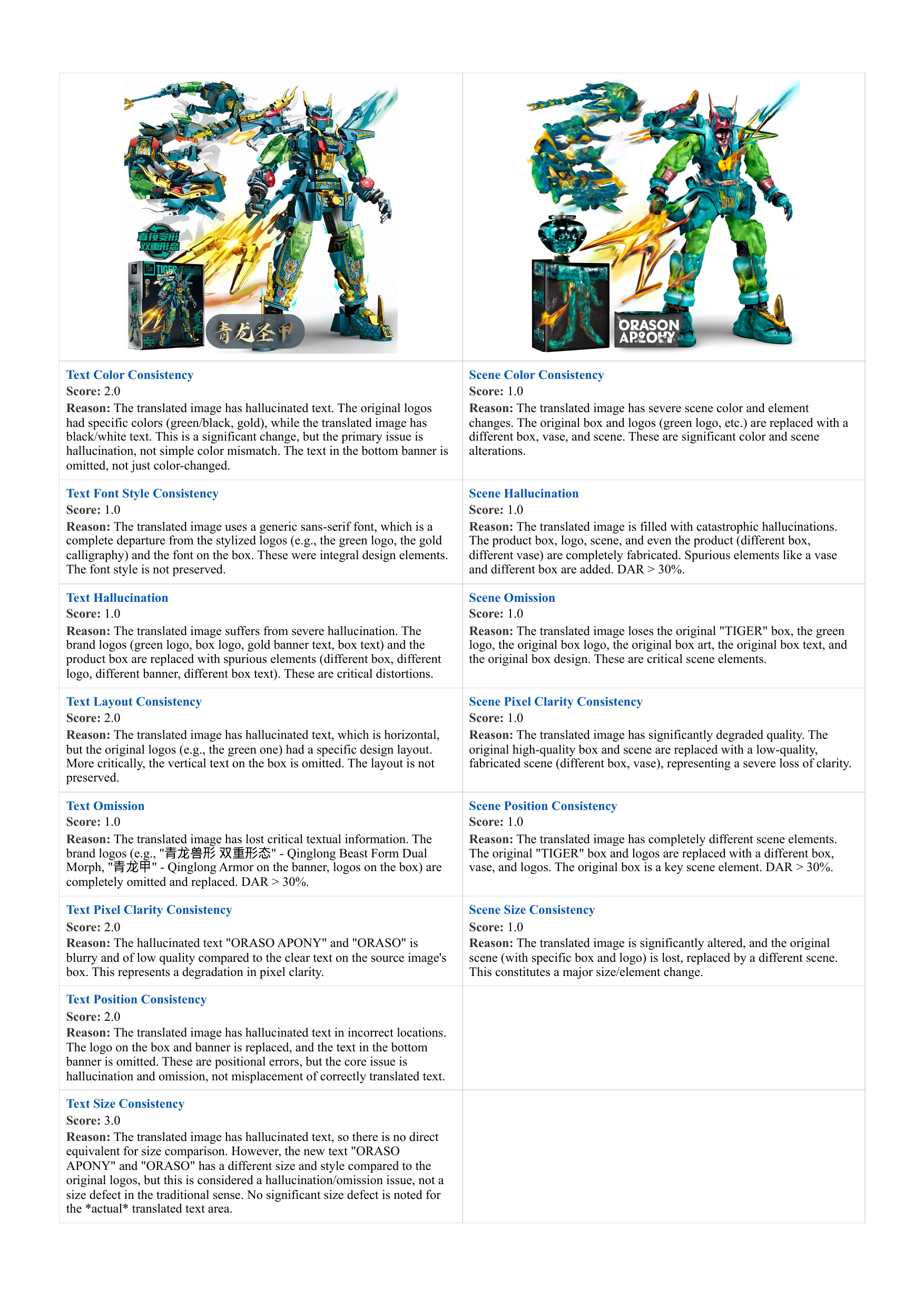}
  \vspace{-1.8cm}
  \caption{E-commerce evaluation example illustrating both large-area hallucinations and fine-grained detail distortions. Left: source image; Right: translated image.}
  \label{fig:eco_example_2}
\end{figure*}

Figure~\ref{fig:eco_example_2} illustrates a common issue with end-to-end MLLM-based systems. When processing images with rich visual details, MLLMs struggle to maintain visual consistency, often resulting in distortions across numerous fine-grained elements. In this evaluation, we observe that Vectra not only identifies the prominent hallucinations and quality defects in the image content, but also performs in-depth inspection of subtle details that are frequently overlooked during manual review---for instance, defects in small text located in less conspicuous regions. Since our 14 dimensions are evaluated independently, the extensive defects in this image are reflected across all scene-related dimensions. For text-related dimensions, while text size consistency is reasonably preserved, all other dimensions exhibit issues, all of which are accurately identified by Vectra.

\begin{figure*}[htbp]
  \centering
  \vspace{-0.3cm}
  \includegraphics[width=\textwidth]{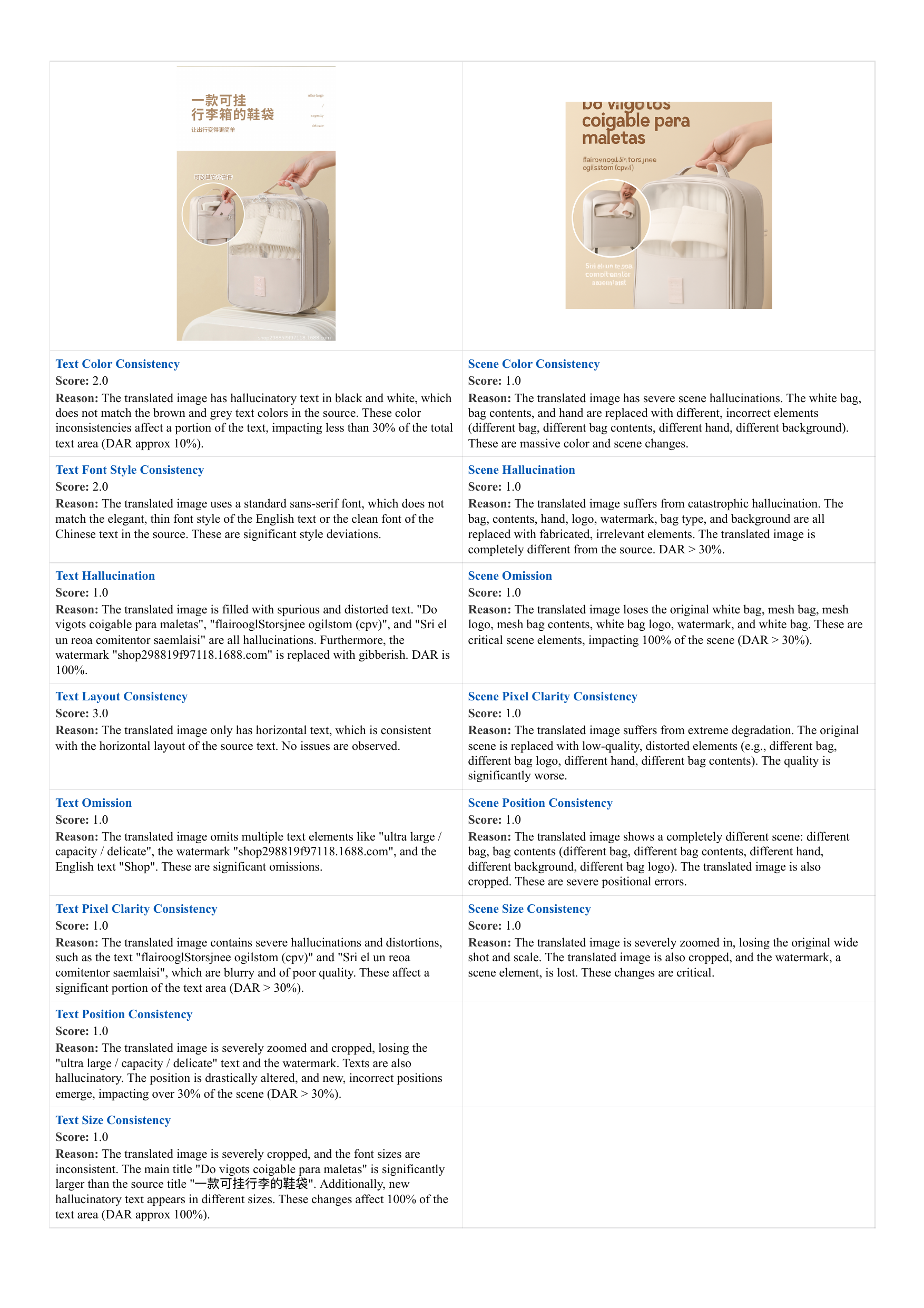}
  \vspace{-1.8cm}
  \caption{E-commerce evaluation example illustrating content truncation due to aspect ratio mismatch. Left: source image; Right: translated image.}
  \label{fig:eco_example_3}
\end{figure*}

%Figure~\ref{fig:eco_example_3} 则反映了另一个问题，很多的数据由于涉及到商品的详情介绍这种下滑形式的页面，图片尺寸存在极大的极大跨度，很多基于MLLM的模型缺乏生成自适应尺度跨度的能力。这会导致其在固定尺寸例如1：1 3:4的比例翻译的较好，对于多样的话的尺寸有时候会导致尺寸的裁剪，从而导致严重的错误
Figure~\ref{fig:eco_example_3} highlights another prevalent issue. Many images originate from scrollable detail pages, resulting in extreme variations in aspect ratios. Many MLLM-based systems lack the capability to generate outputs with adaptive aspect ratios. Consequently, while these models perform reasonably well on fixed aspect ratios (e.g., 1:1 or 3:4), they often resort to cropping when encountering diverse image dimensions, leading to severe content truncation errors.

\subsection{Out-of-Domain Scenarios}
\begin{figure*}[htbp]
  \centering
  \vspace{-0.3cm}
  \includegraphics[width=\textwidth]{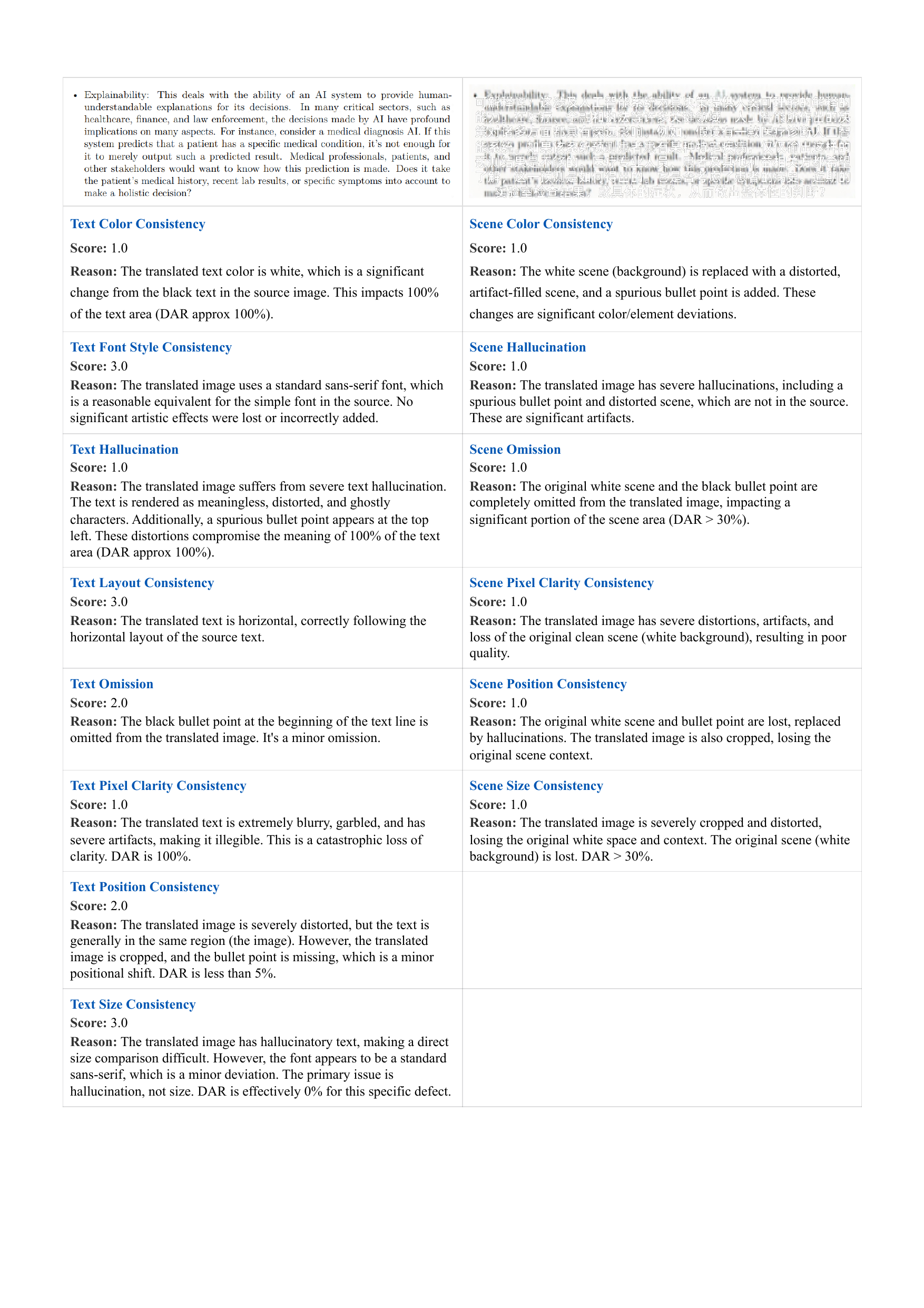}
  \vspace{-4cm}
  \caption{Out-of-domain evaluation example (Document). Left: source image; Right: translated image.}
  \label{fig:eco_example_4}
\end{figure*}

\begin{figure*}[htbp]
  \centering
  \vspace{-0.3cm}
  \includegraphics[width=\textwidth]{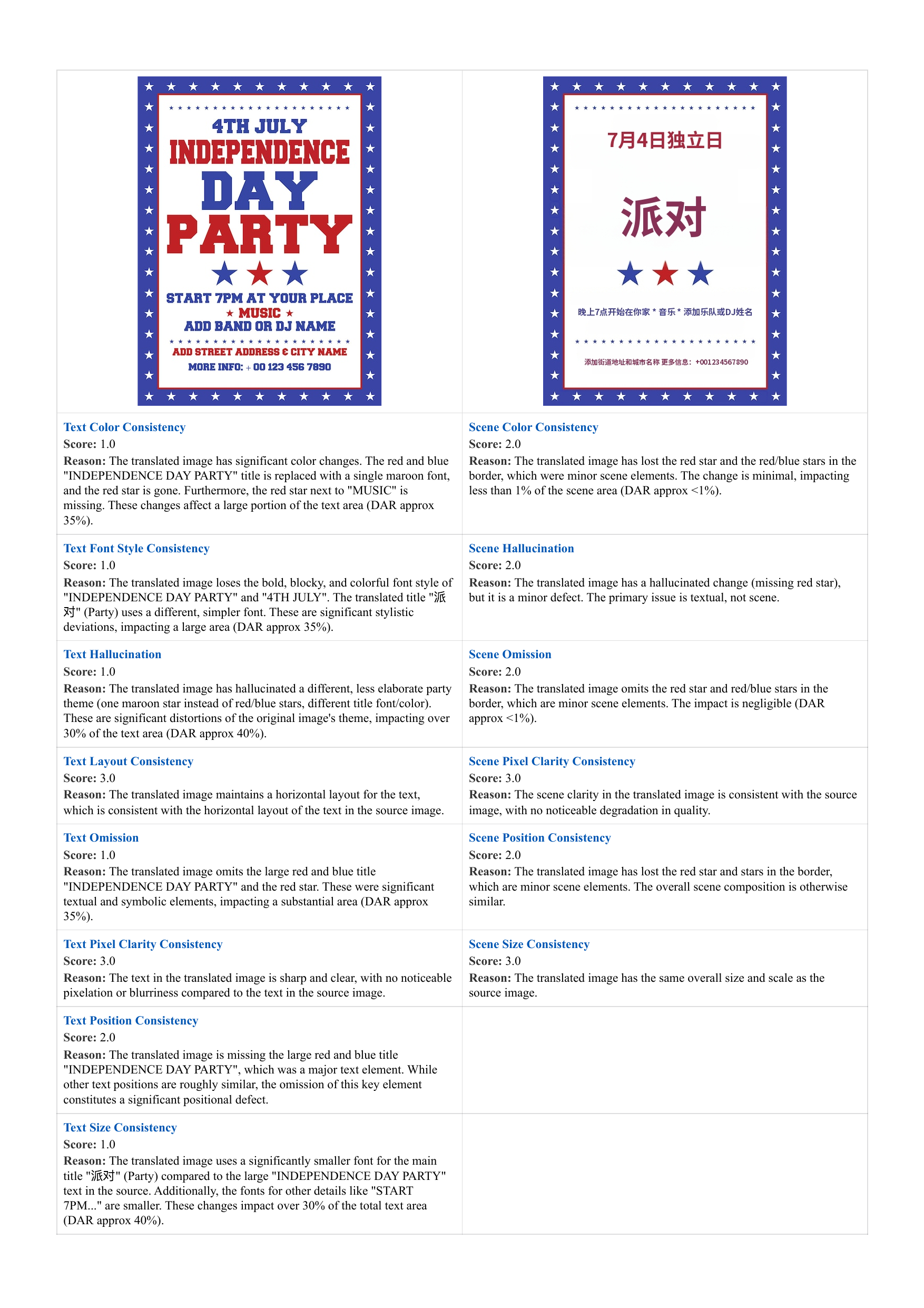}
  \vspace{-1.8cm}
  \caption{Out-of-domain evaluation example (Poster). Left: source image; Right: translated image.}
  \label{fig:eco_example_5}
\end{figure*}

\begin{figure*}[htbp]
  \centering
  \vspace{-0.3cm}
  \includegraphics[width=\textwidth]{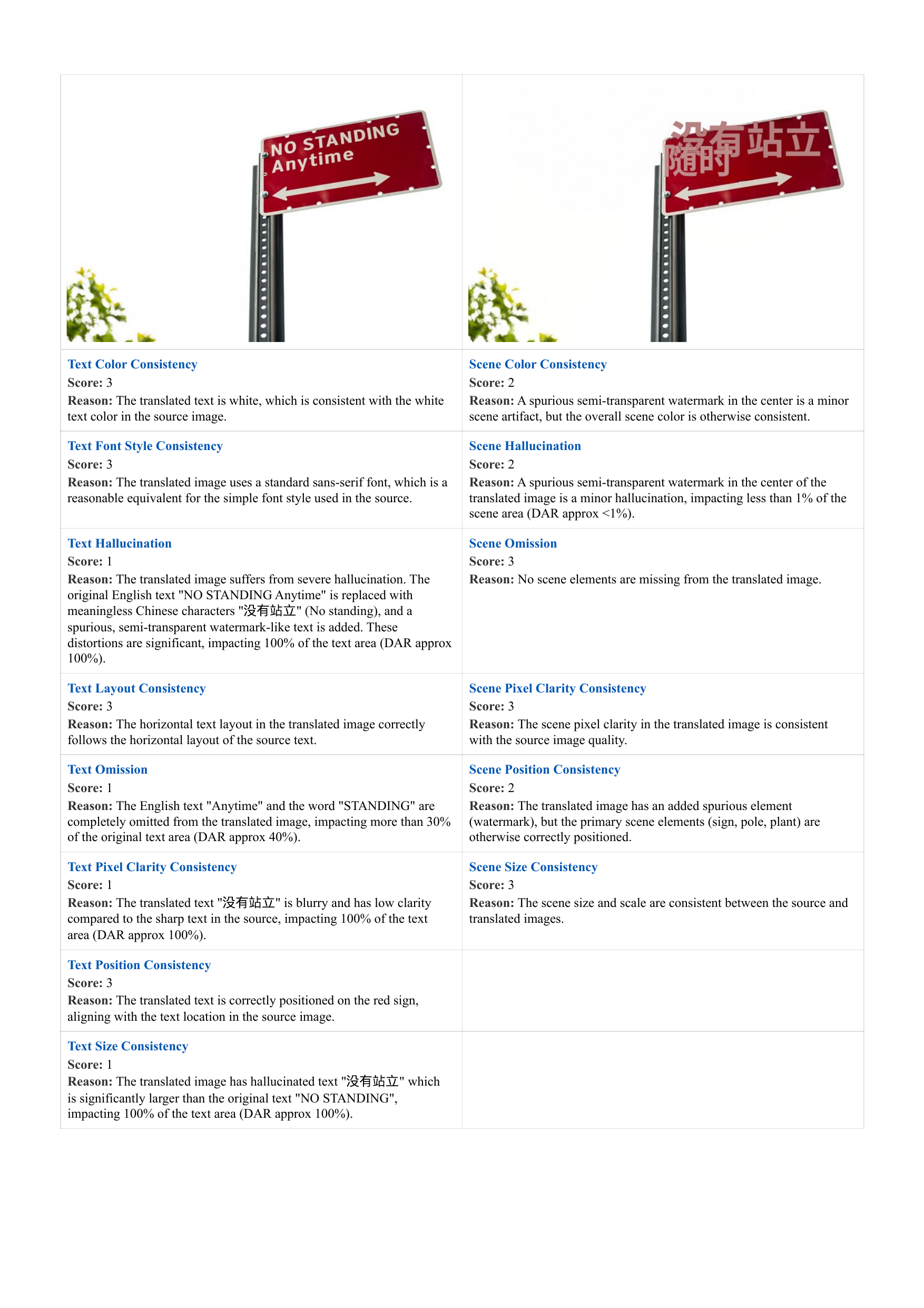}
  \vspace{-3.5cm}
  \caption{Out-of-domain evaluation example (Natural Scene). Left: source image; Right: translated image.}
  \label{fig:eco_example_6}
\end{figure*}

Documents, posters, and natural scenes collectively cover the major application domains of IIMT. We use samples from the MCiT dataset to demonstrate Vectra's evaluation results in these scenarios. Notably, the source images in these domains typically feature clearer text and more structured layouts compared to e-commerce imagery. As illustrated in Figures~\ref{fig:eco_example_4}, \ref{fig:eco_example_5}, and \ref{fig:eco_example_6}, Vectra's evaluation results reveal consistent assessment capabilities across all three scenarios, comparable to its performance on e-commerce data. This demonstrates Vectra's generalization potential.

\end{document}